\documentclass[journal]{IEEEtran}
\usepackage{amsmath}
\usepackage{amssymb}
\usepackage{xcolor}
\usepackage{graphicx}
\usepackage{booktabs}       
\usepackage{subcaption}
\usepackage{subcaption}
\usepackage{float} 
\usepackage{multirow}       
\usepackage[table]{xcolor}  
\usepackage{colortbl}  
\usepackage[ruled,vlined]{algorithm2e}
\usepackage{url}
\definecolor{myblue}{RGB}{230, 245, 255}
\definecolor{mygreen}{RGB}{230, 255, 230}
\definecolor{mypink}{RGB}{255, 240, 245}
\definecolor{myyellow}{RGB}{255, 250, 230}
\definecolor{myorange}{RGB}{255, 240, 220}
\usepackage{caption}
\usepackage{cite}

\ifCLASSINFOpdf
\else
\fi
\begin{document}
%
\title{IoT-Enabled Autonomous Maritime Navigation in Smart Ports: A Curriculum-Guided Shared Policy Learning Framework}

\author{Yuqing Lin,
        Rangya Zhang,
        Kum Fai Yuen
\thanks{Yuqing Lin and Kum Fai Yuen are with the School of Civil and Environmental Engineering, Nanyang Technological University, Singapore (e-mail: yuqing003@e.ntu.edu.sg; kumfai.yuen@ntu.edu.sg).}%
\thanks{Rangya Zhang is with the School of Mechanical and Aerospace Engineering, Nanyang Technological University, Singapore (e-mail: rangya002@e.ntu.edu.sg).}%
\thanks{Corresponding author: Kum Fai Yuen.}}
%
%

\markboth{Journal of \LaTeX\ Class Files,~Vol.~14, No.~8, August~2015}%
{Shell \MakeLowercase{et al.}: Bare Demo of IEEEtran.cls for IEEE Journals}
%



\maketitle

\begin{abstract}
As smart port infrastructures increasingly rely on autonomous maritime devices enabled by the Internet of Things (IoT), ensuring reliable onboard navigation intelligence has become a critical challenge for safe and scalable operations in congested waterways. This paper investigates onboard autonomous navigation for such IoT devices under partial observability and dense traffic conditions. A curriculum-guided reinforcement learning framework with a shared recurrent policy is developed to enhance temporal reasoning, deployment scalability, and robustness of edge-level decision-making. Centralized training is adopted as an offline design-time strategy, while all navigation actions are executed fully onboard, consistent with IoT edge intelligence paradigms. Extensive simulations in multiple realistic port environments demonstrate that the proposed approach improves navigation reliability, collision avoidance, and training stability compared with standard baseline methods, and generalizes effectively to previously unseen high-density scenarios. The results indicate that curriculum-guided shared learning provides a practical solution for scalable deployment of IoT-enabled autonomous maritime devices in smart port operations.

\end{abstract}


%
\IEEEpeerreviewmaketitle

\section{Introduction and Related Work}

\IEEEPARstart{T}{he} rapid digitalization of port infrastructures has accelerated the deployment of IoT-enabled autonomous surface vehicles for inspection, logistics, and traffic support in modern smart ports. Major global hubs such as Singapore, Los Angeles, and Rotterdam have actively invested in data-driven port management, IoT-based vessel monitoring, and autonomous navigation trials. Representative initiatives include Singapore’s autonomous navigation programs integrated with the national Vessel Traffic Information System, Los Angeles’ data-driven berth optimization and vessel traffic management platforms, and Rotterdam’s IoT-enabled vessel monitoring and autonomous vessel testing facilities \cite{mpa2025digitalisation, polastrategic2018, rotterdamai2025}. In such IoT-enabled port environments, autonomous surface vehicles are expected to operate safely and reliably in congested waterways while relying primarily on onboard sensing and decision intelligence.

A fundamental challenge in this context lies in the design of robust onboard navigation intelligence that can function under partial observability, dense traffic, and dynamic environmental disturbances, without assuming continuous communication or global situational awareness. Reliable collision avoidance and regulation-compliant navigation are therefore critical capabilities for the large-scale deployment of IoT-enabled autonomous maritime devices in smart port operations.

\subsection{Onboard Decision Intelligence for Autonomous Maritime IoT Devices}

Deep Reinforcement Learning (DRL) has emerged as a powerful paradigm for onboard sequential decision-making in dynamic and uncertain environments, and has been increasingly adopted in intelligent IoT-enabled autonomous systems for real-time control and decision intelligence~\cite{chang2021survey,wei2020broad}. In maritime autonomous navigation, DRL has been applied to a variety of tasks including trajectory tracking, heading control, disturbance rejection, and collision avoidance~\cite{wen2019optimized, duan2021distributional, zeng2024multi, zhao2025port, hao2023intelligent}. Early applications primarily focused on single-vessel scenarios or simplified interaction settings, demonstrating the feasibility of learning-based controllers as alternatives to classical control methods such as Proportional–Integral–Derivative or Model Predictive Control~\cite{bao2023autonomous, du2021optimized, xue2025improved}. Actor–Critic methods, including A2C and A3C, have further enabled stable learning under stochastic disturbances and partial observability~\cite{wen2019optimized, tang2023path, ye2021a3c}.

As autonomous maritime IoT devices are increasingly deployed in congested and dynamic port environments, reliable onboard decision intelligence becomes critical for safe and scalable operation \cite{liu2020resource}, particularly when navigation must handle complex multi-vessel encounters governed by international regulations such as the International Regulations for Preventing Collisions at Sea (COLREGs)~\cite{niu2023multi, cui2023autonomous}. Many existing approaches embed COLREGs compliance through handcrafted reward terms or heuristic penalties~\cite{guo2021path, wang2021unmanned, lin2024distributional}, which may constrain adaptability when onboard agents are exposed to diverse traffic patterns and environmental conditions. Recent works have explored more structured reward formulations and attention-based architectures to improve robustness and interpretability~\cite{cui2024autonomous, sun2025path}, yet generalization to realistic, high-density port operations remains challenging.

To address partial observability and improve temporal reasoning, DRL-based edge intelligence has been widely explored in IoT-enabled autonomous systems for sequential decision-making under uncertainty~\cite{wu2020adaptive}. In particular, recurrent neural networks such as Long Short-Term Memory (LSTM) and Gated Recurrent Units (GRU) have been integrated into DRL policies to capture temporal dependencies and infer latent environmental states. Recurrent variants of Proximal Policy Optimization (PPO) have demonstrated improved stability and memory-aware decision-making in sequential navigation tasks~\cite{zheng2023partially, rongcai2023autonomous}. In maritime contexts, recurrent PPO-based approaches have shown promising performance in regulation-aware navigation and collision avoidance under dense traffic conditions~\cite{zhang2025adaptive, cui2025collision, sawada2021automatic}. These studies indicate that shared recurrent policy structures can support robust onboard navigation intelligence across diverse deployment scenarios.

\subsection{Curriculum Learning for Robust Onboard Navigation}

Curriculum Learning (CL) improves sample efficiency, training stability, and generalization by gradually increasing task difficulty during learning~\cite{narvekar2020curriculum, portelas2020teacher}. Common strategies include designer-specified curricula~\cite{matiisen2019teacher}, self-paced progression~\cite{ren2018self}, and automatically generated curricula such as reverse curriculum learning~\cite{florensa2017reverse}. CL has been widely adopted in robotics and navigation to mitigate sparse rewards and accelerate convergence in high-dimensional environments.

In intelligent IoT-enabled robotic and autonomous systems, curriculum-based training has shown strong potential in improving robustness and adaptability under real-world uncertainty~\cite{huang2022edge}. In maritime and marine robotics, similar curriculum-based approaches have demonstrated effectiveness in enhancing DRL-driven trajectory planning, obstacle avoidance under environmental disturbances, and regulation-aware navigation behaviors~\cite{marchel2025model, havenstrom2021deep, weng2022reinforcement}. These studies suggest that progressive exposure to increasingly complex navigation scenarios enables onboard agents to acquire safer and more conservative behaviors. However, CL remains underexplored for autonomous surface vessels operating in congested port environments, particularly in conjunction with recurrent policy learning and explicit regulatory constraints. Moreover, existing approaches often rely on heuristic reward shaping rather than embedding navigation regulations directly into the environment and learning process.

\subsection{Contributions and Scope of This Work}

Motivated by the above limitations, this paper focuses on enhancing the onboard decision intelligence of IoT-enabled autonomous surface vehicles through curriculum-guided policy learning. Rather than proposing a new reinforcement learning algorithm, we aim to develop a practical and scalable training framework for robust deployment in smart port environments. The main contributions of this work are summarized as follows:

\begin{itemize}
    \item A shared recurrent policy learning framework for IoT-enabled edge devices, enabling robust onboard decision-making under partial observability and dense traffic conditions without relying on continuous cloud communication.
    \item A curriculum-guided training strategy that progressively exposes the policy to increasing encounter complexity, supporting stable acquisition of safety- and regulation-aware navigation behaviors.
    \item An integrated treatment of safety-critical navigation rules, embedding COLREGs as hard regulatory constraints within the IoT environment and reward formulation to ensure compliant edge-level operations.
    \item Extensive evaluation across multiple geographically distinct smart port scenarios, demonstrating the deployment scalability and generalization of the shared policy under diverse traffic densities and disturbances.
\end{itemize}

The remainder of this paper is organized as follows. Section~\ref{sec:3} introduces the proposed curriculum-guided recurrent policy learning framework, including observation, action, and reward design. Section~\ref{sec:4} presents the simulation environments and experimental setup. Section~\ref{sec:5} reports experimental results and analysis, and Section~\ref{sec:6} concludes the paper with future research directions.

\section{Methodology}
\label{sec:3}
\subsection{Problem Formulation}

We consider onboard autonomous navigation for IoT-enabled autonomous surface vehicles operating in congested port environments under partial observability. Each device relies solely on local sensing and onboard decision intelligence to avoid collisions and reach designated targets while complying with navigation regulations. The navigation task is formulated as a POMDP \cite{spaan2012partially}:
\begin{equation}
\label{eq:tuple}
    \mathcal{P} = \langle \mathcal{S}, \mathcal{A}, \Omega, \mathcal{T}, \mathcal{O}, \mathcal{R}, \gamma \rangle,
\end{equation}
where $\mathcal{S}$ denotes the set of true environmental states, $\mathcal{A}$ the set of possible actions, $\Omega$ the set of possible observations, $\mathcal{T}(s'|s,a)$ the transition function, $\mathcal{O}(o|s,a)$ the observation function, $\mathcal{R}(s,a)$ the reward function, and $\gamma$ the discount factor.

The objective is to learn a stochastic policy $\pi_{\theta}(a|o)$ that maps partial observations to control actions and maximizes the expected discounted return \cite{kurniawati2022partially}:
\begin{equation}
\label{eq:expected}
    \mathbb{E}_{\pi_\theta} \left[ \sum_{t=0}^{\infty} \gamma^t \mathcal{R}(s_t, a_t) \right],
\end{equation}
where $\pi_{\theta}(a|o)$ denotes the policy distribution over actions given observation $o$, $\mathcal{R}(s_t, a_t)$ is the reward obtained at time $t$, and $\gamma \in [0,1)$ balances short- and long-term rewards. The expectation is taken over trajectories generated by following $\pi_{\theta}$.

In realistic port operations, autonomous maritime IoT devices operate under uncertainty and partial observability due to limited sensor range, occlusions, limited communication bandwidth, and environmental disturbances such as currents and wind \cite{ding2023risk, zheng2023partially}. Each device relies on onboard sensors (e.g., radar, LiDAR, proximity sensors) without access to the full global state; real-time synchronization is typically hindered by limited communication bandwidth in remote or congested IoT environments. This motivates the POMDP formulation as interaction patterns must be inferred from local observations.

To improve robustness under partial observability, we adopt a shared recurrent policy learning approach, in which a single policy network is reused across identical devices deployed in shared environments. Reuse of policies enables efficient model deployment at scale while maintaining fully onboard execution. Recurrent LSTM units are integrated into the policy to capture temporal dependencies and infer latent environmental states from observation histories.

\subsection{Shared Recurrent PPO Architecture}
To enable robust onboard collision avoidance for IoT-enabled autonomous surface vehicles operating in congested port environments, we adopt PPO as the underlying learning framework due to its stability and effectiveness in continuous control tasks. A shared recurrent policy is trained offline using experience collected from multiple autonomous surface vehicles operating in parallel simulation environments, and subsequently deployed onboard each device for independent execution. This shared-policy design facilitates efficient model reuse across identical platforms and improves training stability under diverse initial conditions and encounter scenarios. CL is employed during training to progressively increase task complexity, enhancing learning efficiency and policy generalization.

To address partial observability arising from limited sensor range and occlusions caused by surrounding vessels and obstacles, LSTM units are embedded within both the actor and critic networks. This recurrent design enables the policy to capture temporal dependencies and infer latent environmental states from observation histories, which is consistent with the POMDP formulation of the navigation task.

\begin{figure}[t]
    \centering
    \includegraphics[width=\linewidth]{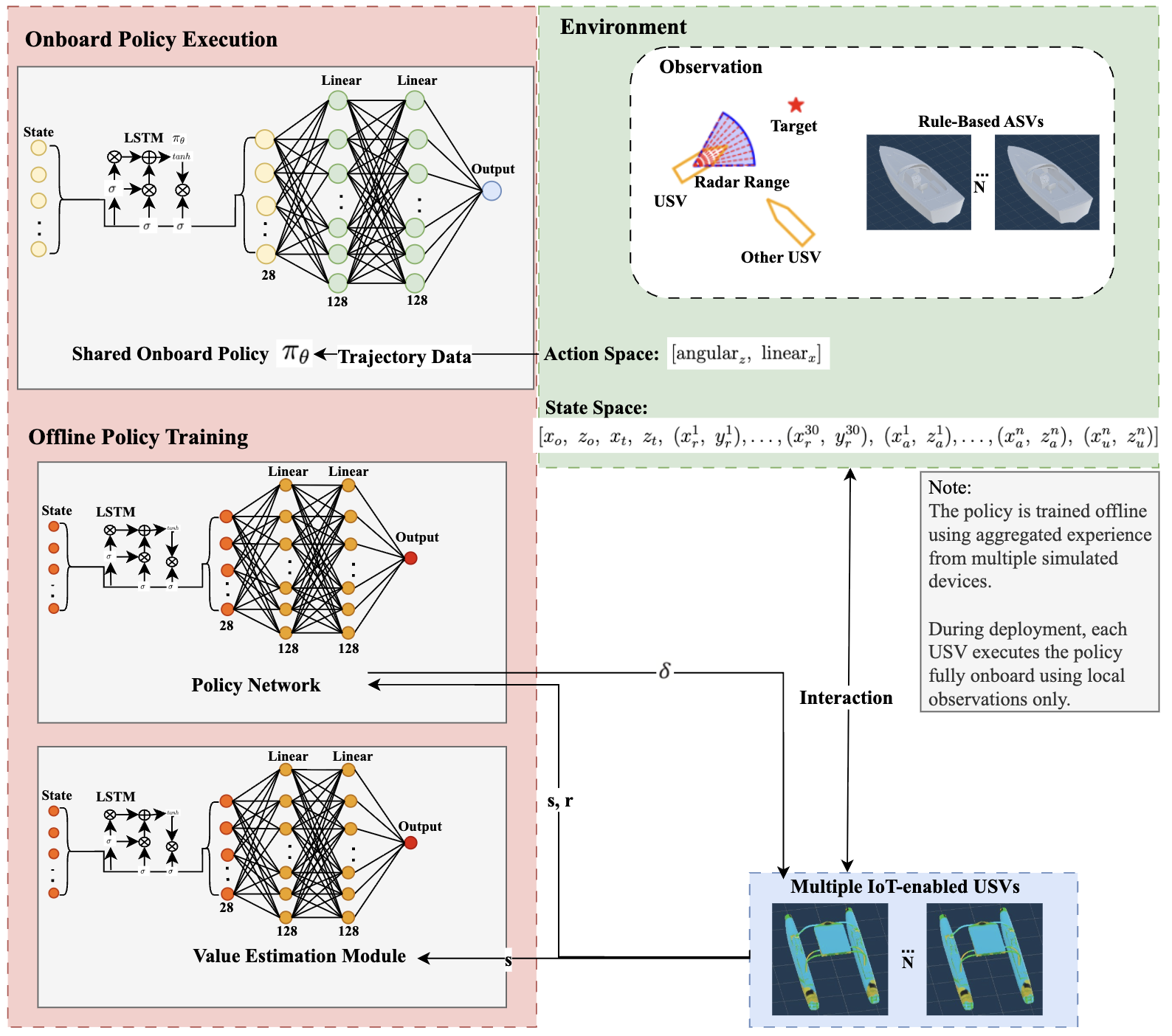}
    \caption{Framework of the proposed onboard recurrent policy learning approach for IoT-enabled autonomous surface vehicles. The policy is trained offline and executed fully onboard using local observations.}
    \label{fig:ppo_architecture}
\end{figure}

The overall architecture, illustrated in Fig.~\ref{fig:ppo_architecture}, integrates recurrent policy learning with fully onboard execution to handle dynamic and partially observable maritime environments. During deployment, each autonomous surface vehicle operates solely based on local observations, including radar-based detection of static and dynamic obstacles (buoys and other vessels), relative coordinates to the navigation target, normalized distances to environmental boundaries, relative positions of nearby vessels, and ego-motion variables. The continuous action space consists of commanded yaw rate and forward surge velocity.

Observations are processed sequentially through an LSTM layer to maintain an internal hidden state over time. The LSTM output is passed through fully connected layers to generate the parameters of a Gaussian action distribution, from which continuous control actions are sampled. This architecture enables temporal reasoning and improves decision robustness under uncertainty. During training, experience trajectories from multiple simulated vehicles are aggregated to update the shared policy parameters. Both actor and critic networks share the same recurrent backbone, consisting of an LSTM layer followed by two linear layers, with the actor outputting Gaussian policy parameters and the critic estimating the state value function. PPO with a clipped surrogate objective is adopted to ensure stable policy updates, while Generalized Advantage Estimation is used to reduce variance in advantage computation.

By combining a shared recurrent policy with curriculum-guided training, the proposed framework supports robust onboard decision-making under partial observability and dense traffic conditions, while remaining scalable for deployment across multiple IoT-enabled autonomous surface vehicles operating in realistic port environments.

\subsection{Observation and Action Space}

Onboard autonomous navigation for IoT-enabled autonomous surface vehicles requires informative yet partial perception of surrounding traffic and obstacles under realistic sensor constraints. The observation space is designed to capture both geometric layout and dynamic environmental information while accounting for limitations such as finite radar range and occlusion. Radar-based perception discretizes the sensing field into angular sectors, with each sector encoding the normalized distance and bearing to the nearest detected object. All quantities are expressed in the body-fixed frame to ensure rotational invariance and normalized according to sensor and velocity limits. Temporal consistency across observations is preserved through the use of an LSTM-based policy, which is essential for decision-making under POMDP conditions.

\begin{figure}[t]
    \centering
    \includegraphics[width=0.85\linewidth]{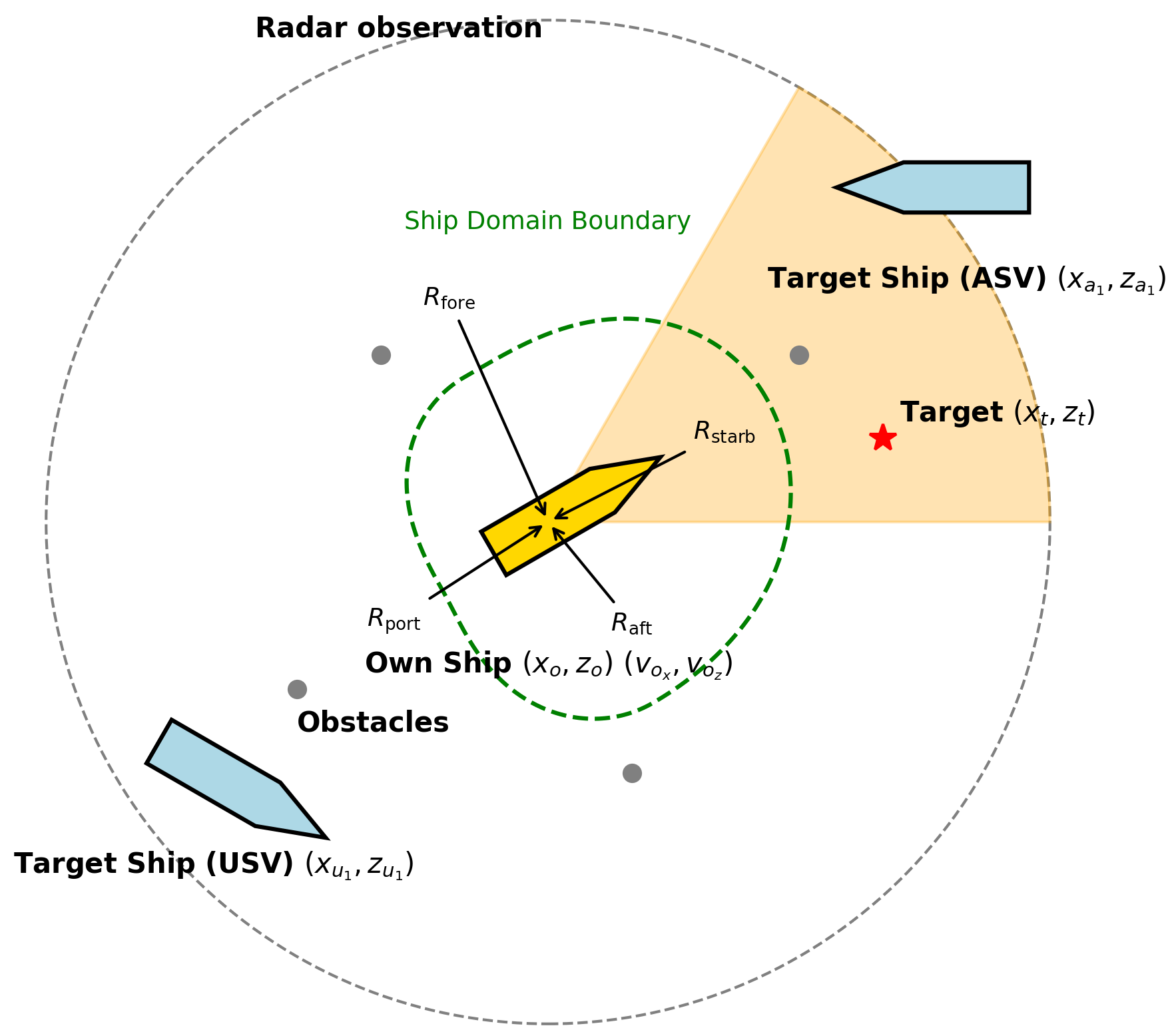}
    \caption{Ship perception model and onboard observation structure.}
    \label{fig:shipdomain}
\end{figure}

To structure onboard situational awareness, an anisotropic elliptical ship domain with direction-dependent radii \(R_{\text{fore}}, R_{\text{aft}}, R_{\text{port}}, R_{\text{starb}}\) is adopted to reflect varying clearance priorities in different directions (Fig.~\ref{fig:shipdomain})~\cite{szlapczynski2017review, wang2010intelligent}. This domain serves as a perception and safety abstraction that adapts with vessel orientation and supports regulation-aware risk assessment compatible with COLREGs-based reward design. Within this representation, each device perceives its own kinematic state, navigation goal, surrounding traffic vessels with heterogeneous control behaviors, and static obstacles. The resulting observation vector is defined as:
\begin{equation}
\label{eq:observation}
\begin{aligned}
\big[\, 
&\underbrace{x_o, z_o, v_{ox}, v_{oz}}_{\text{Own ship}},\ 
\underbrace{x_t, z_t}_{\text{Goal}},\ 
\underbrace{x_{a_1}, z_{a_1}, x_{u_1}, z_{u_1}, \dots}_{\text{Surrounding vessels}}, \\
&\underbrace{d_1, \theta_1, d_2, \theta_2, \dots}_{\text{Radar obstacles}} 
\,\big]
\end{aligned}
\end{equation}
where all positions and velocities are expressed relative to the own ship, and \((d_i, \theta_i)\) denote radar-derived obstacle ranges and bearings.

\begin{figure}[t]
    \centering
    \includegraphics[width=0.9\linewidth]{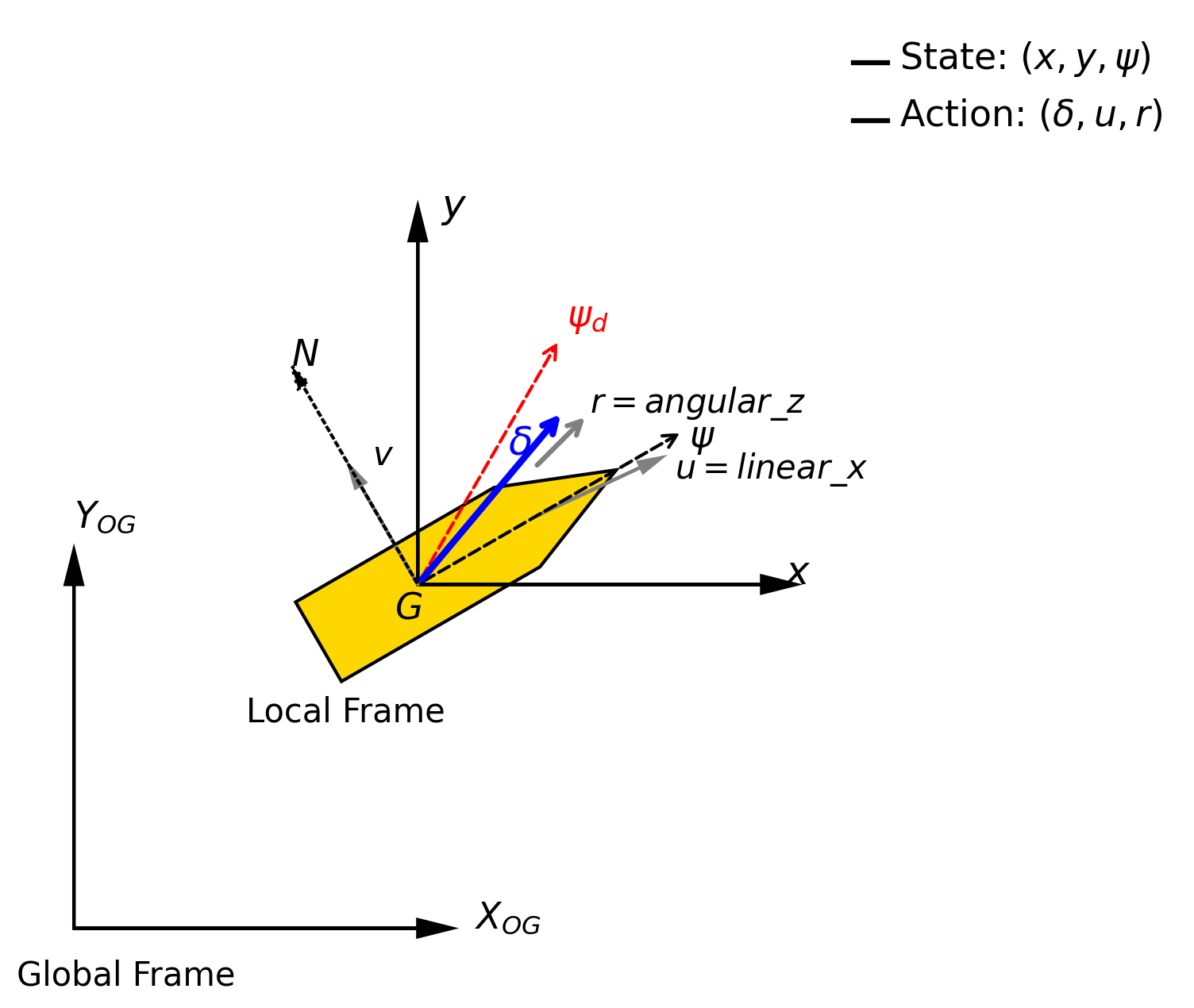}
    \caption{3-DOF kinematic model and onboard action space.}
    \label{fig:shipdynamic}
\end{figure}

For motion control, a simplified three-degree-of-freedom (3-DOF) kinematic model in the horizontal plane is adopted, including surge \(u\), sway \(v\), and yaw rate \(r\), while neglecting vertical-plane dynamics to reduce modeling complexity~\cite{fossen2011handbook}. The body-fixed reference frame is centered at the vessel’s center of mass \(G\), with heading angle \(\psi\) defined relative to the global frame (Fig.~\ref{fig:shipdynamic}). The continuous two-dimensional action space is defined as:
\begin{equation}
\label{eq:action}
[\text{angular}_z,\ \text{linear}_x] = [r,\ u]
\end{equation}
where \(\text{linear}_x\) corresponds to the commanded surge velocity and \(\text{angular}_z\) to the yaw rate. This high-level action abstraction decouples decision-making from low-level actuation details such as propeller thrust and rudder dynamics, enabling efficient onboard policy learning and robust generalization under environmental disturbances. The resulting control interface supports a range of navigation behaviors, including trajectory tracking, maneuvering in constrained waterways, and regulation-compliant collision avoidance.

\subsection{Reward Shaping Strategy}
\label{sec:reward_shaping}

To support robust onboard decision-making for IoT-enabled autonomous surface vehicles operating in congested port environments, we design a composite reward that balances navigation efficiency, safety preservation, and regulation-aware risk mitigation. Rather than encoding explicit maneuvering rules, the reward function provides continuous guidance signals derived from geometric relationships, proximity risk, and encounter context, enabling effective learning under partial observability.

The overall reward consists of four dense components, heading alignment $r_{\text{yaw}}$, distance progress $r_{\text{dist}}$, regulation-aware interaction $r_{\text{reg}}$, and proximity-based safety penalty $r_{\text{prox}}$, supplemented by sparse terminal rewards.

\paragraph{Heading alignment}
This term encourages progress toward the navigation target by aligning the vessel’s heading with the desired bearing:
\begin{equation}
r_{\text{yaw}} = \cos(\psi - \psi_d),
\end{equation}
where $\psi$ denotes the current heading and $\psi_d$ the bearing toward the target.

\paragraph{Distance progress}
To promote efficient navigation, distance progress is rewarded as:
\begin{equation}
r_{\text{dist}} = \lambda_{\text{dist}} \cdot (\text{Dist}_{t-1} - \text{Dist}_t),
\end{equation}
where $\lambda_{\text{dist}}$ is a scaling coefficient and $\text{Dist}_t$ denotes the normalized distance to the target at time $t$.

\paragraph{Regulation-aware interaction reward}
To incorporate maritime navigation regulations as part of onboard risk assessment~\cite{naeem2012colregs, kuwata2013safe}, a regulation-aware interaction reward $r_{\text{reg}}$ is activated when surrounding traffic vessels enter the anisotropic ship domain of the own vessel (Fig.~\ref{fig:shipdomain}). The ship domain is parameterized by direction-dependent radii $R_{\text{fore}}$, $R_{\text{aft}}$, $R_{\text{port}}$, and $R_{\text{starb}}$, providing a continuous representation of encounter proximity and relative geometry.

\begin{figure}[t]
    \centering
    \includegraphics[width=0.9\linewidth]{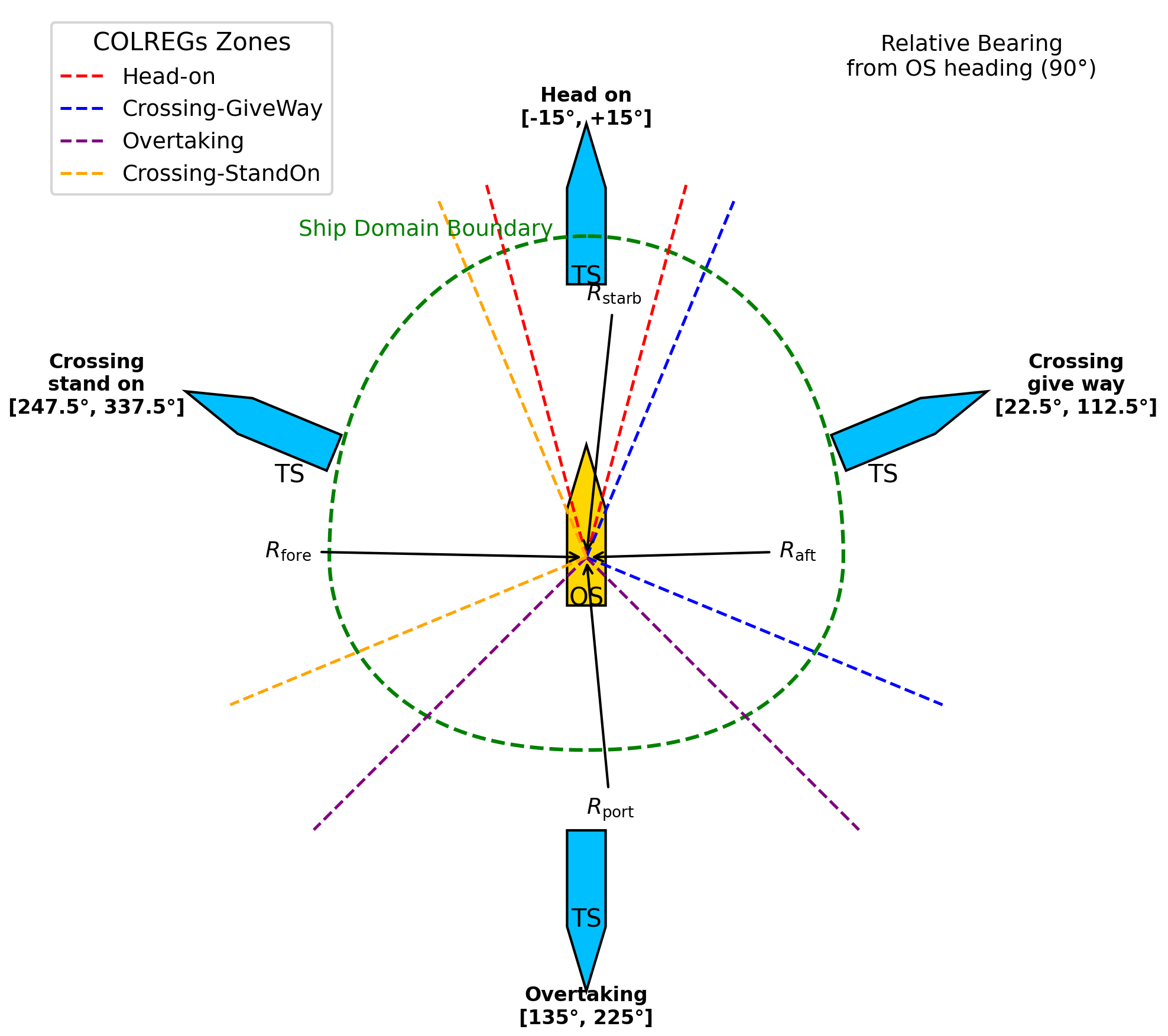}
    \caption{Regulation-aware encounter sectors in the anisotropic ship domain.}
    \label{fig:colregs_reward}
\end{figure}

Encounter contexts are categorized based on relative bearing and domain location (Fig.~\ref{fig:colregs_reward}), and each context is associated with a signed shaping signal that reflects relative collision risk and navigational responsibility:
\begin{itemize}
    \item Head-on encounters emphasize lateral separation.
    \item Crossing encounters differentiate between yielding and stand-on responsibilities.
    \item Overtaking encounters emphasize longitudinal clearance.
    \item Parallel navigation emphasizes safe lateral spacing.
\end{itemize}
The total regulation-aware reward aggregates contributions from all surrounding traffic vessels:
\begin{equation}
\label{eq:colregs_reward}
\begin{aligned}
r_{\text{reg}}
= \sum_{j \in \mathcal{T}} \bigl(
&\mathcal{T}_{\text{head-on}}(j) f_{\text{head-on}} + 
\mathcal{T}_{\text{cross-give}}(j) f_{\text{cross-give}} \\
&+ \mathcal{T}_{\text{overtake}}(j) f_{\text{overtake}} +
\mathcal{T}_{\text{cross-stand}}(j) f_{\text{cross-stand}} \\
&+ \mathcal{T}_{\text{parallel}}(j) f_{\text{parallel}}
\bigr),
\end{aligned}
\end{equation}
where $\mathcal{T}$ denotes the set of surrounding traffic vessels detected within sensing range.

\paragraph{Proximity-based safety penalty}
To discourage unsafe proximity, a continuous penalty is applied when the distance $d$ to a nearby vessel falls below a minimum safety threshold $d_{\text{min}}$:
\begin{equation}
r_{\text{prox}} = 
\begin{cases}
-\lambda_{\text{prox}} \left(1 - \frac{d}{d_{\text{min}}} \right), & d < d_{\text{min}}, \\
0, & \text{otherwise}.
\end{cases}
\end{equation}

\paragraph{Terminal rewards}
Sparse terminal rewards reflect task completion and safety outcomes:
\begin{itemize}
    \item Successful arrival at the navigation target: $+1.0$.
    \item Collision or violation of navigable boundaries: $-1.0$.
    \item Episode timeout without reaching the target: $-0.6$.
\end{itemize}

\paragraph{Final reward}
The per-step reward is computed as:
\begin{equation}
r_t = r_{\text{yaw}} + r_{\text{dist}} + r_{\text{reg}} + r_{\text{prox}}.
\end{equation}

This reward formulation integrates geometric guidance, safety margins, and regulation-aware risk cues into a unified onboard shaping strategy, enabling sample-efficient learning and robust navigation under dense traffic and partial observability without relying on explicit rule execution.

\subsection{CL Strategy}

Training autonomous surface vehicles for reliable onboard navigation in congested port environments with static obstacles and dynamic surrounding traffic presents challenges in convergence, safety preservation, and generalization. To address these issues, we adopt a hybrid CL strategy that combines designer-specified task staging~\cite{narvekar2020curriculum} with self-paced progression driven by policy performance~\cite{ren2018self}. Task difficulty is gradually increased by adjusting environmental and operational conditions, and policy advancement occurs only when predefined proficiency criteria are satisfied. This structured learning process improves sample efficiency, convergence stability, and robustness for onboard decision-making under partial observability.

Each curriculum stage varies along three key dimensions:

\begin{itemize}
    \item Deployment Scale: Training begins with a single autonomous device operating independently, and progressively increases the number of simultaneously deployed devices. This allows the shared policy to mature under increasing traffic density while preserving fully onboard decision-making without explicit coordination.
    
    \item Traffic Interaction Complexity: Early stages involve simple or limited surrounding traffic. Later stages introduce heterogeneous autonomous surface vehicles executing randomized regulation-consistent behaviors (e.g., head-on, overtaking, crossing, and parallel encounters), increasing encounter diversity and improving robustness to unseen traffic patterns.
    
    \item Environmental Complexity: Static obstacle density, navigational constraints, and environmental disturbances (e.g., wind and current) are gradually increased to better approximate real-world operating conditions and enhance adaptability.
\end{itemize}

Table~\ref{tab:curriculum} summarizes the curriculum structure. Stage~1 (Port of Los Angeles) focuses on isolated interaction learning under a deterministic traffic configuration. Stage~2 (Port of Singapore) introduces moderate traffic variability through randomized encounter scenarios. Stage~3 (Port of Rotterdam) exposes the policy to dense and heterogeneous traffic conditions to promote generalization under realistic operational complexity. Curriculum progression is triggered by either a fixed number of training episodes or performance thresholds (e.g., success rate $\geq 80\%$ over a moving evaluation window).

This staged training strategy enables stable policy development under safety-critical and partially observable conditions, while providing a systematic framework for evaluating onboard navigation performance across increasingly realistic and complex port environments.

\begin{table*}[t]
\centering
\setlength{\tabcolsep}{3pt}
\renewcommand{\arraystretch}{1.05}

\begin{tabular}{
    c c p{1.2cm} c c |
    c c p{1.2cm} c c |
    c c p{1.2cm} c c
}
\toprule
\multicolumn{5}{c|}{Stage 1: Port of Los Angeles} 
& \multicolumn{5}{c|}{Stage 2: Port of Singapore}
& \multicolumn{5}{c}{Stage 3: Port of Rotterdam} \\
\cmidrule(lr){1-5} \cmidrule(lr){6-10} \cmidrule(lr){11-15}
USVs & ASVs & Encounter Types & Obstacles & Weather 
& USVs & ASVs & Encounter Types & Obstacles & Weather
& USVs & ASVs & Encounter Types & Obstacles & Weather \\
\midrule
1 & 0 & None & Sparse & None 
  & 2 & 2 & Head-on / Crossing / Overtaking / Parallel (random) & Moderate & None 
  & 4+ & 4+ & Mixed encounters (randomized) & Dense & Wind + Waves \\
  
\addlinespace[1mm]
1 & 1 & Head-on / Crossing / Overtaking / Parallel (fixed) & Sparse & None 
  & 2 & 4 & Head-on / Crossing / Overtaking / Parallel (random) & Dense & Waves 
  & -- & -- & -- & -- & -- \\
\bottomrule
\end{tabular}

\vspace{0.5em}
\caption{Curriculum stages with progressively increasing difficulty and randomized ASV encounters per episode. 
Stage~1 corresponds to the Port of Los Angeles (open and sparse traffic), Stage~2 to the Port of Singapore (semi-structured and high-traffic), and Stage~3 to the Port of Rotterdam (hybrid, complex transitions).}
\label{tab:curriculum}
\end{table*}

\section{Simulation Environment}
\label{sec:4}
\subsection{Platform Overview}
The simulation framework is implemented using Unity ML-Agents, which provides a flexible 3D physics-based environment for training and evaluating onboard decision policies. The Unity PhysX engine enables realistic vessel dynamics, collision handling, and motion modeling, while high-fidelity rendering supports accurate scenario representation.

Each autonomous surface device is equipped with simulated maritime sensing capabilities, including radar-like proximity sensing for obstacle and traffic detection, lidar-style range measurements for spatial awareness, and relative goal-position inputs for navigation. These complementary sensing modalities allow the onboard policy to perceive local surroundings and make informed navigation decisions under dynamic traffic and environmental uncertainty. Vessel motion is modeled using a simplified differential thrust scheme, where the policy outputs continuous commands for angular velocity and forward propulsion that are mapped to left and right thruster forces within physical constraints. The vessel model incorporates inertia, turning delay, and velocity-dependent dynamics, making navigation and collision avoidance non-trivial. Boundary interactions and buoyancy effects are handled by the underlying physics engine.

A shared recurrent PPO policy with LSTM memory is used to enable sequential decision-making under partial observability. The shared policy allows consistent deployment across multiple autonomous devices while preserving fully onboard execution without requiring explicit coordination, supporting robust navigation across diverse operational scenarios.

\subsection{Scenario Design}
\label{sec:scenario design}

To enhance generalization and real-world robustness for onboard autonomous navigation, we adopt a staged curriculum learning (CL) framework across three representative port environments of increasing operational complexity: Los Angeles (open water, low traffic), Singapore Pasir Panjang (semi-structured, dense traffic), and Rotterdam (hybrid layout with complex transitions). Task difficulty is progressively increased across stages, allowing the onboard policy to mature in handling collision avoidance, regulation-aware navigation, and adaptive behavior under diverse maritime conditions. Progression between stages is governed by either a fixed episode budget or performance thresholds.

\begin{figure}[t]
    \centering
    \begin{subfigure}[b]{0.48\linewidth}
        \centering
        \includegraphics[width=\linewidth]{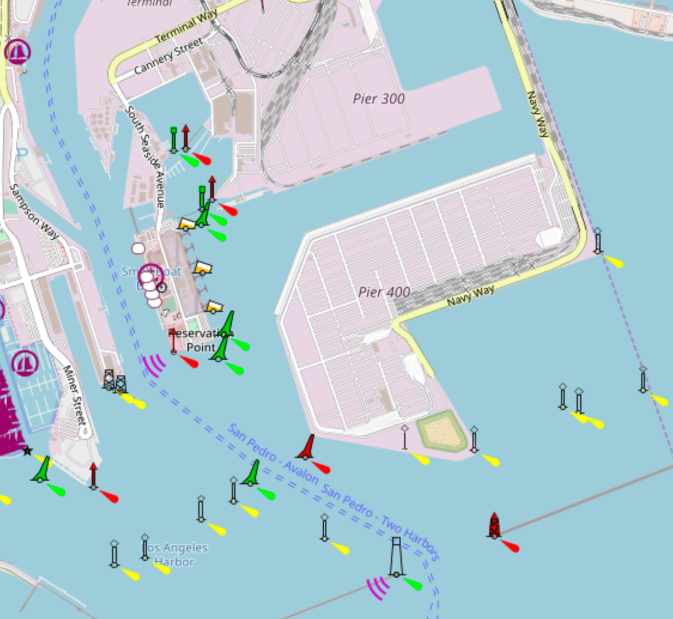}
        \caption{LA port waters from OpenSeaMap.}
        \label{fig:lareal}
    \end{subfigure}
    \hfill
    \begin{subfigure}[b]{0.48\linewidth}
        \centering
        \includegraphics[width=\linewidth]{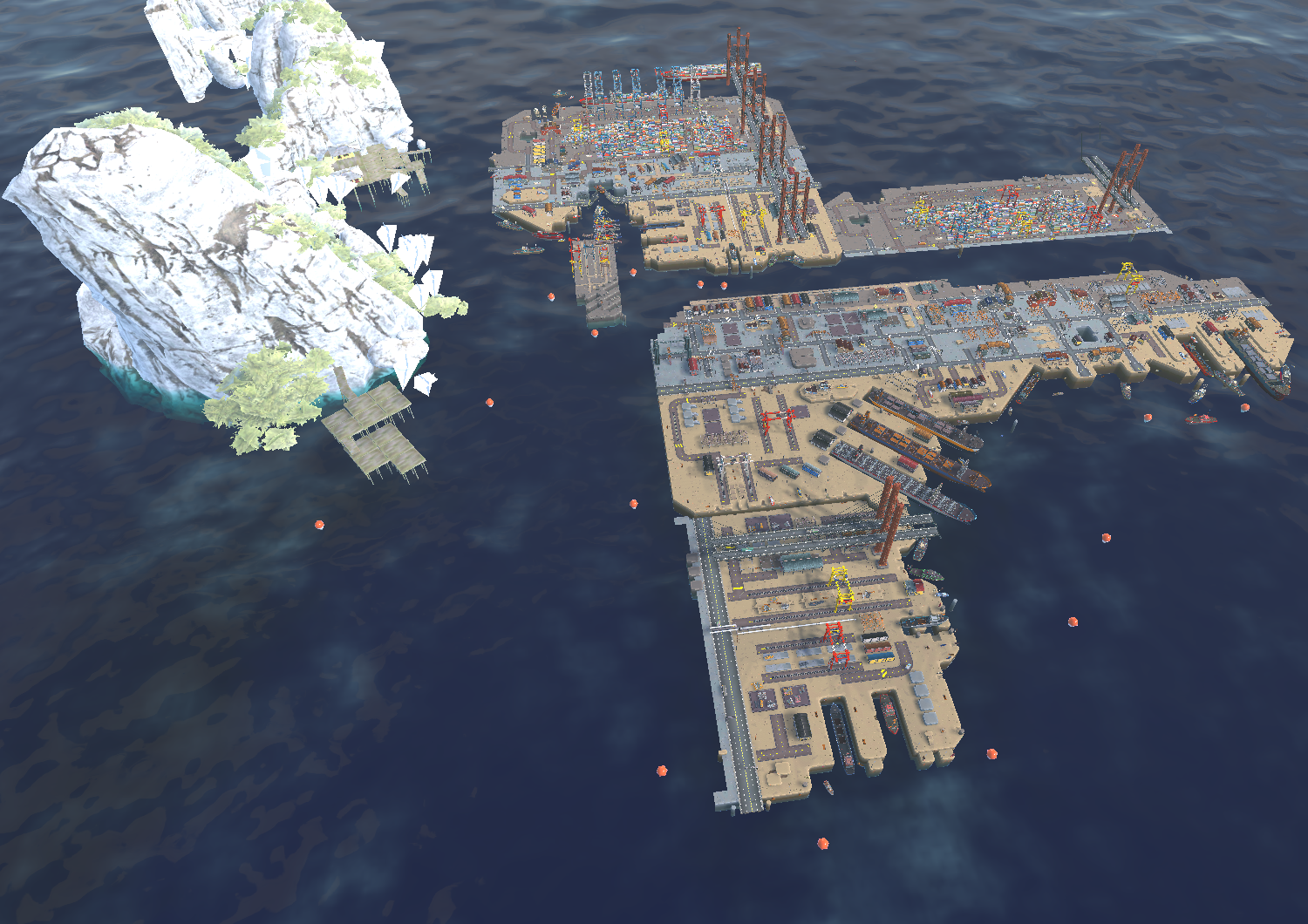}
        \caption{Simulation training environment based on LA port layout.}
        \label{fig:lasimpleenv}
    \end{subfigure}
    \caption{Mapping real-world maritime structure to simulation at the port of Los Angeles.}
    \label{fig:lascenario}
\end{figure}

\paragraph{Stage~1: LA Port (Open, Low-Traffic Waters)}  
A simplified offshore environment inspired by the Port of Los Angeles with low traffic density and large maneuvering space (Fig.~\ref{fig:lascenario}). A single surrounding traffic vessel executes a deterministic regulation-consistent encounter (e.g., head-on, crossing, overtaking, or parallel), enabling the autonomous device to develop stable navigation and energy-efficient maneuvering in a controlled setting.

\begin{figure}[t]
    \centering
    \begin{subfigure}[b]{0.48\linewidth}
        \centering
        \includegraphics[width=\linewidth]{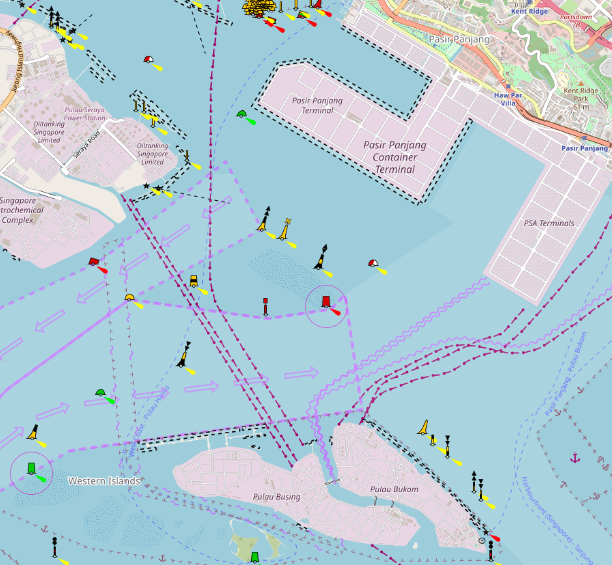}
        \caption{Singapore port waters from OpenSeaMap.}
        \label{fig:sgreal}
    \end{subfigure}
    \hfill
    \begin{subfigure}[b]{0.48\linewidth}
        \centering
        \includegraphics[width=\linewidth]{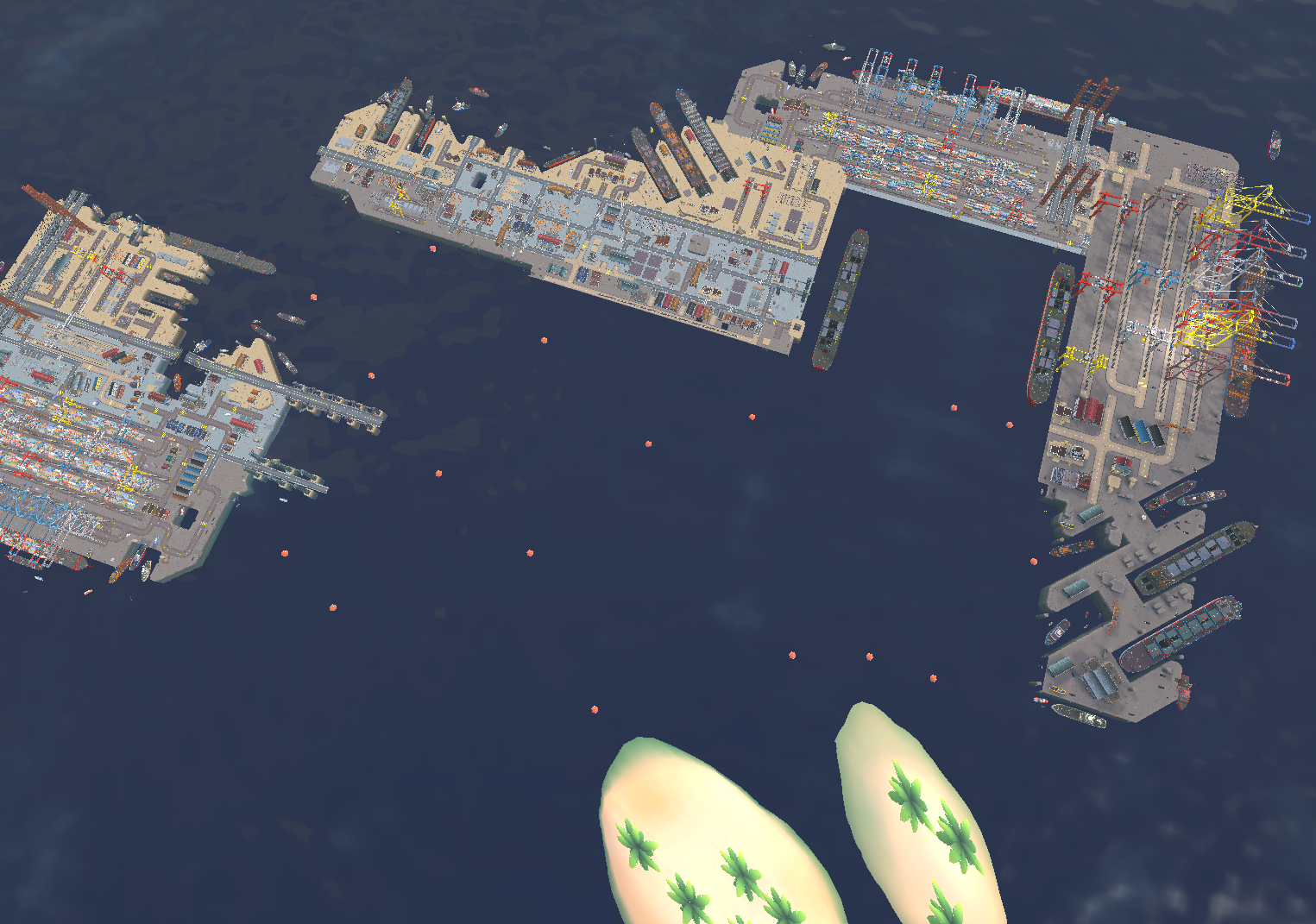}
        \caption{Simulation training environment based on Singapore port layout.}
        \label{fig:sgsimpleenv}
    \end{subfigure}
    \caption{Mapping real-world maritime structure to simulation at the Port of Singapore.}
    \label{fig:sgscenario}
\end{figure}

\footnotetext{\url{https://map.openseamap.org}}
\paragraph{Stage~2: Singapore Port (Semi-Structured, Dense Traffic)}  
A constrained fairway environment based on the Pasir Panjang Container Terminal (Fig.~\ref{fig:sgscenario}), featuring narrow channels and increased traffic density. Two surrounding traffic vessels with randomized regulation-consistent behaviors are introduced, exposing the onboard policy to more complex interaction patterns under congested navigational constraints.

\begin{figure}[t]
    \centering
    \begin{subfigure}[b]{0.48\linewidth}
        \centering
        \includegraphics[width=\linewidth]{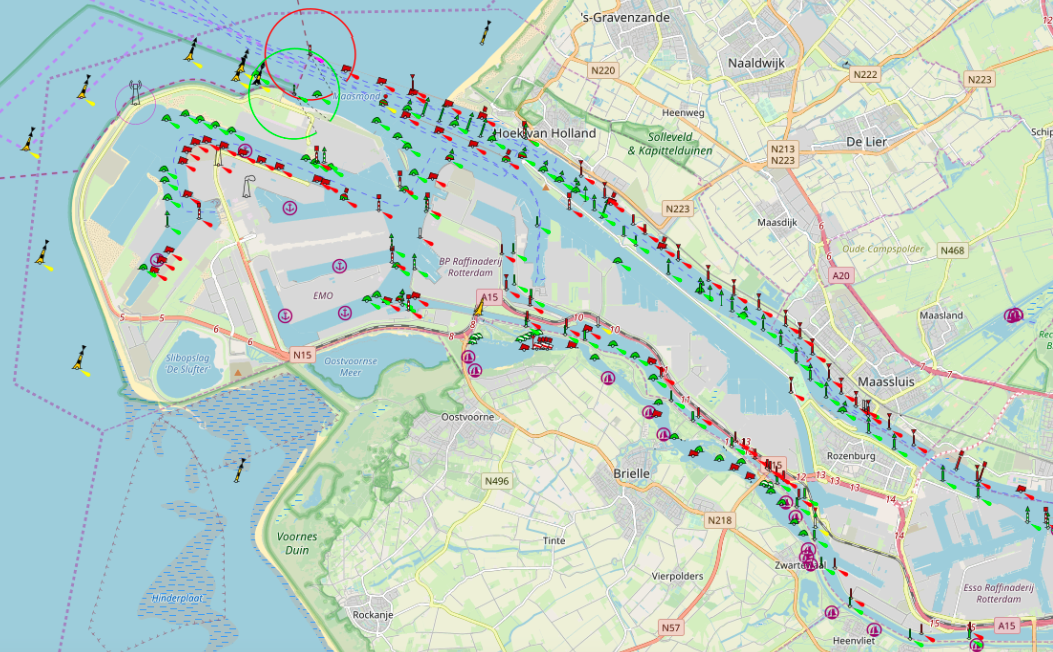}
        \caption{Rotterdam port waters from OpenSeaMap.}
        \label{fig:rreal}
    \end{subfigure}
    \hfill
    \begin{subfigure}[b]{0.48\linewidth}
        \centering
        \includegraphics[width=\linewidth]{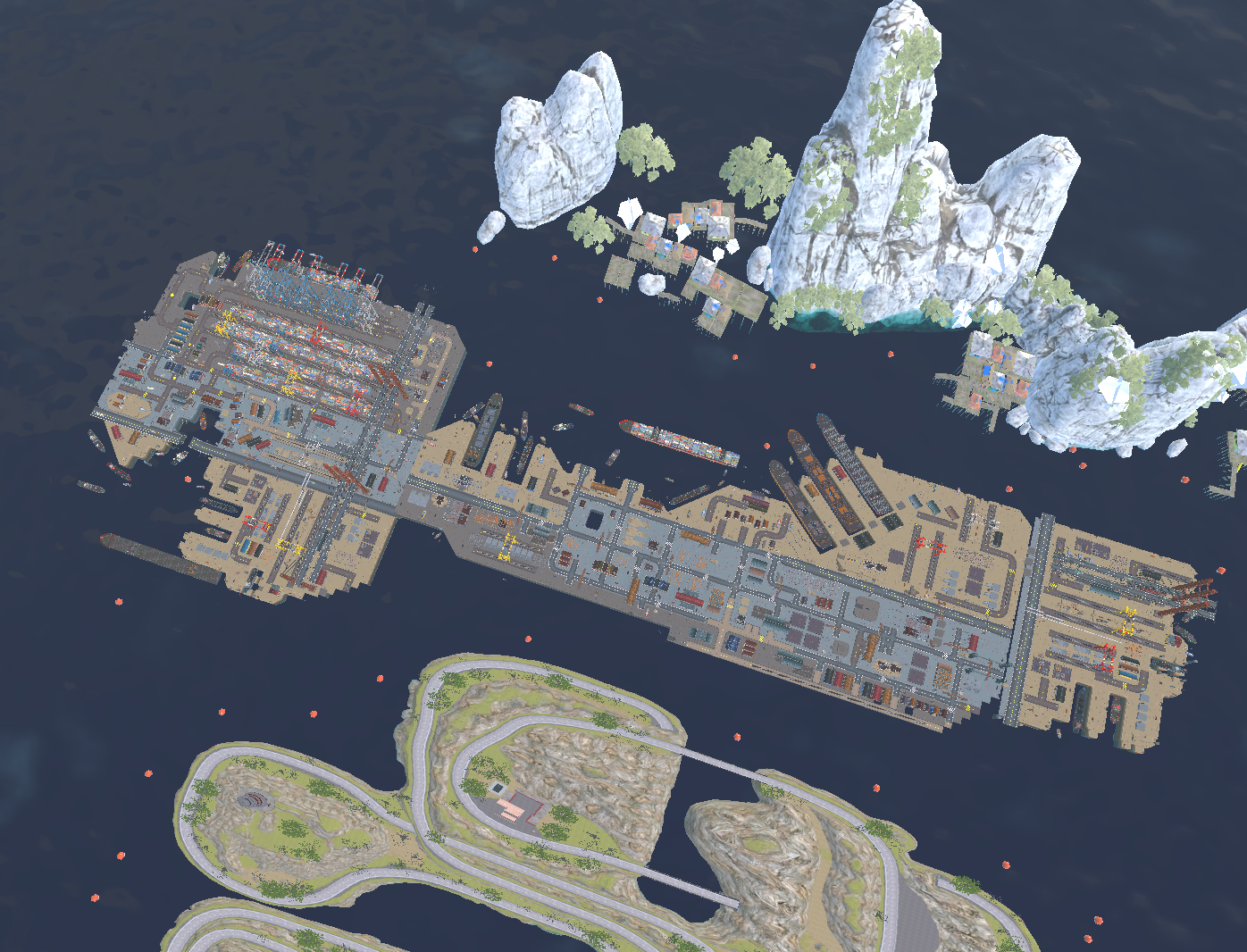}
        \caption{Simulation training environment based on Rotterdam port layout.}
        \label{fig:rsimpleenv}
    \end{subfigure}
    \caption{Mapping real-world maritime structure to simulation at the port of Rotterdam.}
    \label{fig:rscenario}
\end{figure}

\paragraph{Stage~3: Rotterdam Port (Hybrid, Complex Transitions)}  
A hybrid environment modeled after the Port of Rotterdam (Fig.~\ref{fig:rscenario}) containing both narrow channels and open basins. Multiple autonomous surface devices operate concurrently within dense and heterogeneous traffic, where surrounding vessels follow randomized regulation-consistent encounter patterns. This stage introduces complex traffic merging, environmental disturbances, and long-horizon decision challenges, promoting robust policy generalization under realistic operational conditions.

Table~\ref{tab:curriculum} summarizes the curriculum structure. Across stages, scripted surrounding vessels generate diverse and realistic encounter conditions, enabling the onboard policy to progressively adapt from low-risk scenarios to dense and uncertain traffic environments. This geography-informed curriculum design supports scalable and transferable onboard navigation suitable for deployment in diverse real-world port settings.

\subsection{Training Process and Parameters }

\begin{figure}[t]
    \centering
    \includegraphics[width=0.95\linewidth]{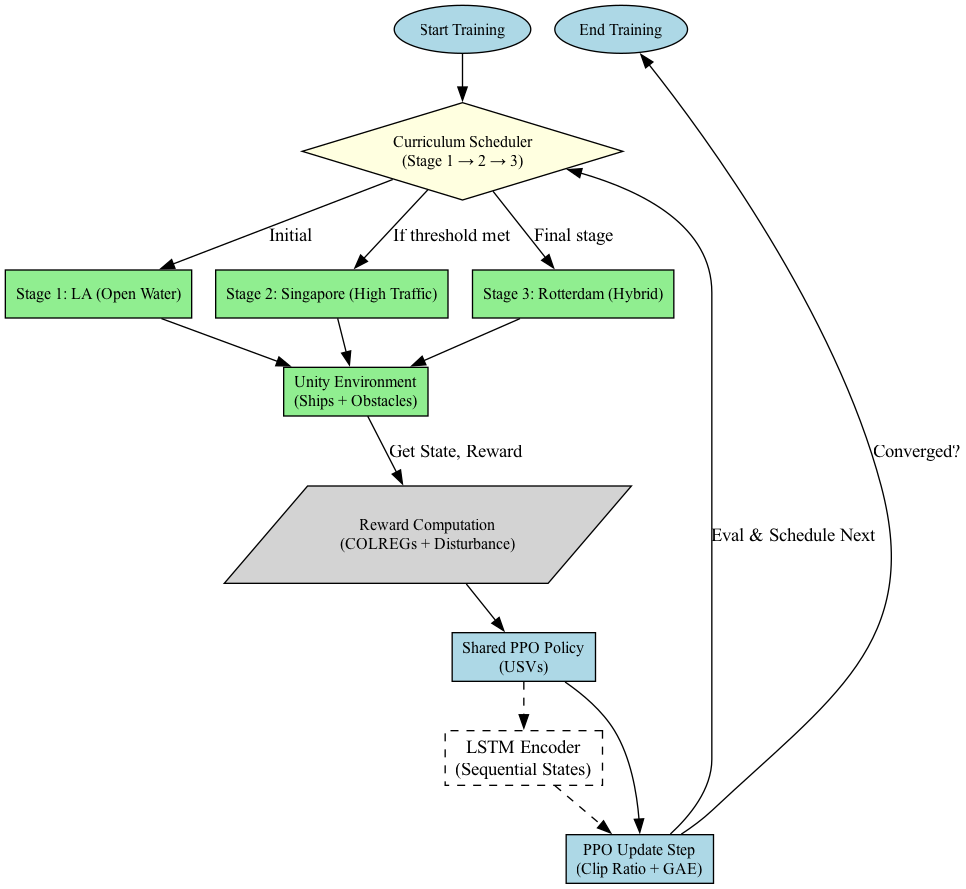}
    \caption{Training process of curriculum-guided shared recurrent PPO.}
    \label{fig:training}
\end{figure}

The training framework in Figure~\ref{fig:training} employs a hybrid CL strategy to incrementally expose agents to progressively challenging multi-vessel navigation scenarios. Three stages of increasing complexity are used: (1) open, low-traffic waters in Los Angeles, (2) semi-structured, high-traffic port approaches in Singapore, and (3) hybrid port layouts in Rotterdam with narrow waterways, high vessel density, complex traffic transitions, and environmental disturbances (wind and waves). A curriculum scheduler monitors success rates and advances to the next stage once proficiency thresholds are met, ensuring stable skill acquisition.

At the beginning of each episode, a curriculum scheduler selects the training environment within a Unity-based maritime simulator containing static obstacles, dynamic surrounding vessels following regulation-consistent trajectories, and stochastic environmental disturbances. Each autonomous surface device receives multimodal observations, including relative goal position, radar-like proximity sensing, and lidar-based range measurements. Observations are normalized and temporally stacked to form a sequential state representation for onboard decision-making.

The proposed policy follows a shared recurrent PPO architecture, consisting of a multilayer perceptron encoder and an LSTM module to capture temporal dependencies under partial observability. The actor head outputs Gaussian-distributed steering and propulsion commands, while the critic head estimates state values for policy optimization. A single shared policy is trained and then consistently deployed across multiple autonomous devices, enabling scalable onboard execution without requiring explicit coordination.

Training proceeds iteratively through parallel trajectory collection, reward evaluation using a composite shaping function (navigation efficiency, regulation-aware safety, and collision avoidance), and PPO updates based on generalized advantage estimation and a clipped surrogate objective. Performance indicators such as navigation success rate, collision frequency, and regulation-compliance score determine curriculum progression, allowing the policy to gradually adapt to increasingly complex operational conditions. Training is executed asynchronously using the PPO trainer in Unity ML-Agents, and key hyperparameters are summarized in Table~\ref{tab:ppo_hyperparams}.

\begin{table}[ht]
\centering
\begin{tabular}{ll}
\hline
Hyperparameter & Value \\
\hline
Batch size & 2048 steps \\
Buffer size & 40960 \\
Learning rate & $3 \times 10^{-4}$ \\
Discount factor ($\gamma$) & 0.99 \\
GAE lambda & 0.95 \\
PPO clip range & 0.2 \\
LSTM hidden size & 128 \\
Number of environments & 8 parallel simulations \\
Total training steps & 5 million \\
\hline
\end{tabular}
\caption{Key PPO hyperparameters used during training with Unity ML-Agents.}
\label{tab:ppo_hyperparams}
\end{table}

\section{Experiments and Results}
\label{sec:5}
\subsection{Algorithm Comparison}

All experiments are run on a workstation with an NVIDIA RTX 4080 GPU and Intel i7 CPU. Each training run lasts 5M interaction steps (6–8 hours), using Unity ML-Agents with 8 parallel environments. The curriculum progresses through three port scenarios: Stage~1 (Los Angeles) features open-water navigation with static obstacles; Stage~2 (Singapore) introduces congested harbor traffic with multiple COLREGs-compliant vessels; Stage~3 (Rotterdam) adds hybrid layouts with wind, waves, dense traffic, and complex docking. Agents advance to the next stage upon achieving an 80\% success rate over 100 evaluation episodes.

We compare the proposed shared recurrent PPO framework against three representative learning paradigms: (1) DDPG~\cite{lillicrap2015continuous}, a deterministic off-policy method without curriculum or recurrence; (2) SAC~\cite{haarnoja2018soft}, an entropy-regularized off-policy method without memory; and (3) standard PPO, trained without curriculum learning or recurrence. These baselines evaluate how different learning paradigms influence onboard navigation reliability and scalability for IoT-enabled autonomous devices.

To ensure fair comparison, all algorithms share identical observation, action, reward formulation, and network capacity. Differences arise only from learning paradigm (on/off-policy), recurrent memory, and curriculum scheduling. The objective is not to outperform algorithmic state-of-the-art, but to validate the effectiveness of a deployable shared onboard policy framework under realistic maritime IoT conditions.

\begin{figure}[t]
    \centering
    \includegraphics[width=0.8\linewidth]{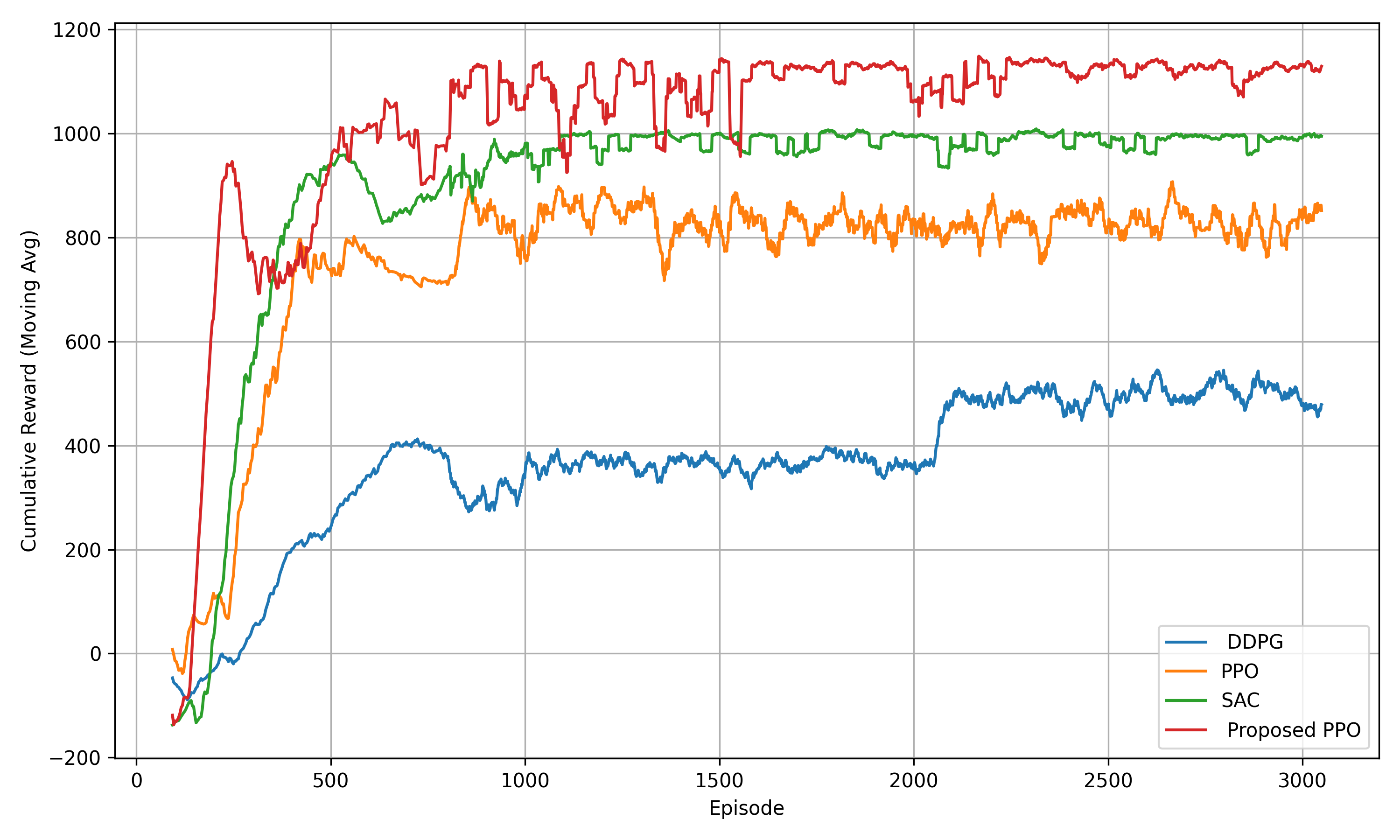}
    \caption{Convergence comparison of onboard navigation policies under curriculum training.}
    \label{fig:totalreward}
\end{figure}

As shown in Fig.~\ref{fig:totalreward}, the proposed PPO demonstrates more stable convergence and higher final performance under complex traffic conditions, validating the effectiveness of shared recurrent policy learning for onboard navigation. SAC learns rapidly in early stages but plateaus in complex scenarios due to the absence of memory and curriculum exposure. Standard PPO converges slower and achieves lower final performance, while DDPG exhibits unstable learning under partial observability.

Table~\ref{tab:algo_comparison} confirms these trends across metrics. In Stage~1, all methods perform well, though SR-PPO achieves the highest success (96.2\%) and lowest collisions (2.1\%). In Stage~2, traffic complexity widens the gap: SR-PPO maintains strong performance with 90.5\% success and 4.0\% collisions, slightly ahead of SAC and clearly more stable than PPO and DDPG. In Stage~3, under dense interactions and disturbances, SR-PPO continues to demonstrate robust behavior with 85.6\% success and 6.3\% collisions. SAC and PPO degrade under increasing complexity, while DDPG shows the weakest performance. Generalization scores follow a similar trend, with SR-PPO consistently achieving strong performance across stages.

\begin{table*}[t]
\centering
\small
\setlength{\tabcolsep}{4pt}
\renewcommand{\arraystretch}{1.1}
\begin{tabular}{l|p{2cm}|p{2cm}|p{2cm}|p{2cm}|p{2cm}}
\toprule
Algorithm & Success Rate (\%) & Collision Rate (\%) & Min. Passing Dist. (m) & Episode Duration (steps) & Generalization Score (\%) \\
\midrule
\multicolumn{6}{c}{Stage 1: Los Angeles (Open Water)} \\
\midrule
Proposed PPO & 96.2 ± 2.1 & 2.1 ± 1.0 & 2.2 ± 0.4 & 82.4 ± 6.7 & 91.3 ± 3.2 \\
SAC          & 94.1 ± 2.8 & 2.3 ± 1.4 & 2.2 ± 0.4 & 80.5 ± 7.1 & 89.4 ± 4.2 \\
DDPG         & 87.9 ± 4.5 & 8.9 ± 3.1 & 2.5 ± 0.7 & 102.3 ± 10.1 & 76.2 ± 6.3 \\
PPO          & 90.6 ± 3.8 & 6.4 ± 2.9 & 2.8 ± 0.5 & 93.7 ± 7.9 & 81.1 ± 4.6 \\
\midrule
\multicolumn{6}{c}{Stage 2: Singapore (High Traffic)} \\
\midrule
Proposed PPO & 90.5 ± 3.3 & 4.0 ± 1.8 & 2.8 ± 0.3 & 96.7 ± 7.5 & 88.9 ± 3.9 \\
SAC          & 87.3 ± 4.1 & 4.2 ± 3.1 & 2.7 ± 0.5 & 100.2 ± 8.6 & 80.6 ± 5.5 \\
DDPG         & 68.5 ± 5.7 & 21.3 ± 6.4 & 2.1 ± 0.6 & 126.1 ± 11.8 & 60.9 ± 7.3 \\
PPO          & 83.7 ± 4.6 & 12.1 ± 4.3 & 2.5 ± 0.5 & 108.5 ± 8.6 & 74.4 ± 6.1 \\
\midrule
\multicolumn{6}{c}{Stage 3: Rotterdam (Hybrid Conditions)} \\
\midrule
Proposed PPO & 85.6 ± 4.0 & 6.3 ± 2.2 & 3.5 ± 0.4 & 109.9 ± 9.1 & 85.2 ± 4.7 \\
SAC          & 78.2 ± 6.2 & 18.1 ± 5.4 & 2.5 ± 0.6 & 121.8 ± 10.4 & 69.8 ± 7.9 \\
DDPG         & 55.7 ± 7.1 & 34.5 ± 7.3 & 1.9 ± 0.5 & 139.6 ± 12.7 & 49.3 ± 8.4 \\
PPO          & 70.2 ± 5.7 & 16.8 ± 4.7 & 2.4 ± 0.5 & 123.1 ± 9.4 & 66.7 ± 7.4 \\
\bottomrule
\end{tabular}
\vspace{0.5em}
\caption{Performance comparison of onboard navigation policies across three IoT-enabled port environments.}
\label{tab:algo_comparison}
\end{table*}

Therefore, curriculum learning and recurrent policy design improve the robustness, safety, and reliability of onboard navigation under partial observability. While SAC remains competitive in simpler environments, the absence of recurrence and curriculum limits its performance in complex scenarios. The results highlight the effectiveness of shared onboard policy learning for scalable and reliable autonomous navigation in IoT-enabled maritime environments.

\subsection{Ablation Study Results}

\begin{figure}[t]
    \centering
    \includegraphics[width=\linewidth]{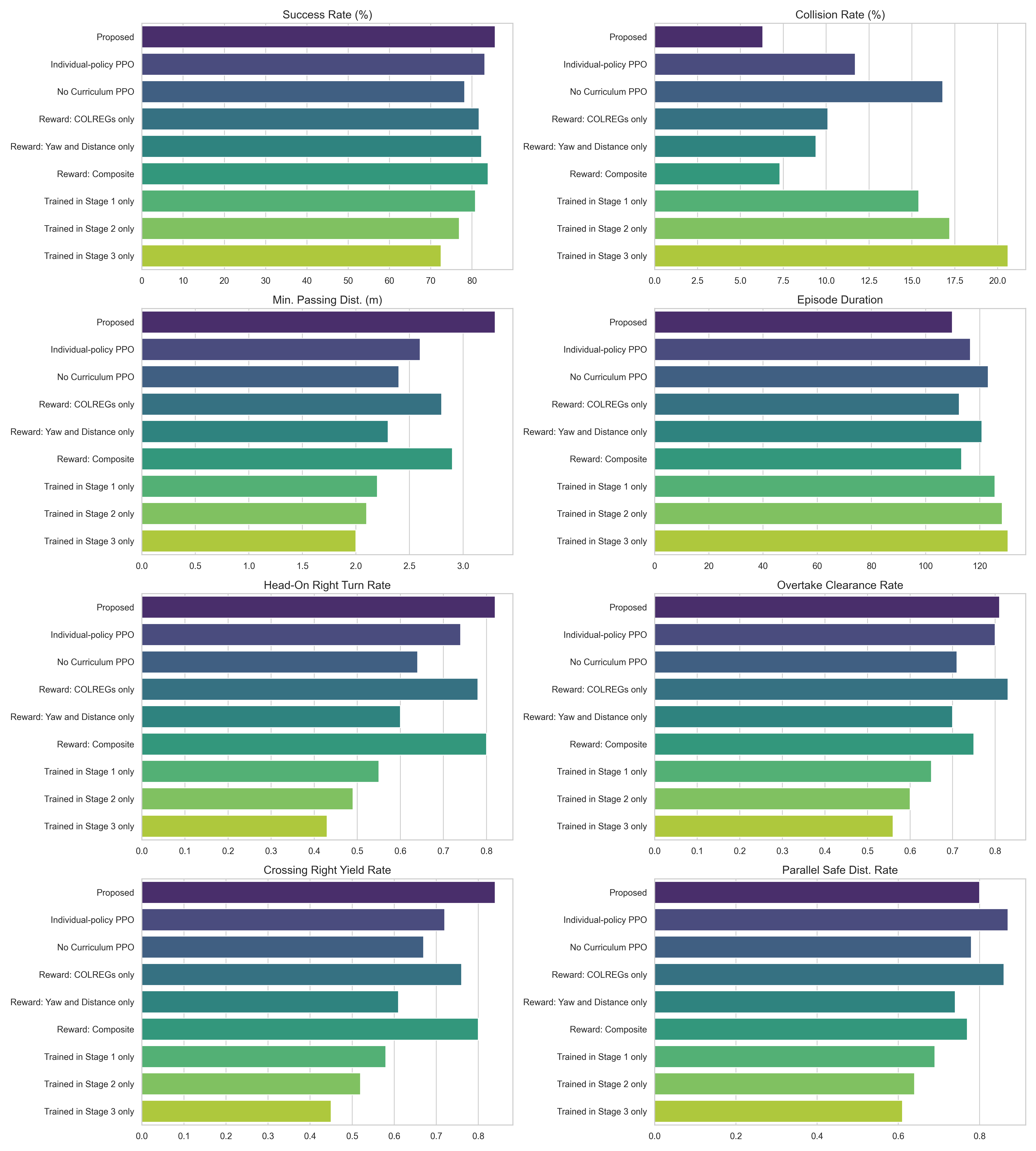}
    \caption{Ablation study of key design choices. Metrics for the first six configurations are averaged in Stage~3. The last three configurations are trained and evaluated independently in Stage~1, 2, or 3 to examine CL benefits.}
    \label{fig:ablation}
\end{figure}

We evaluate the contribution of each core component through ablation studies, selectively removing or modifying elements of the proposed framework. Figure~\ref{fig:ablation} summarizes performance across navigation effectiveness, safety, and regulation compliance. The first six configurations are trained using the full curriculum and evaluated in Stage~3, while the last three are trained and evaluated solely within Stage~1, Stage~2, or Stage~3 to analyze the effect of curriculum progression.

Success Rate (\%): The proposed method achieves the highest overall success rate, improving by approximately 10--15\% compared to Individual-policy PPO and No-Curriculum PPO. Reward-ablation variants (COLREGs-only or Distance-only) show moderate degradation (6--12\%), while Stage~3-only training experiences over 20\% reduction, indicating the importance of shared policy learning and staged curriculum exposure for reliable onboard navigation.

Collision Rate (\%): The proposed model maintains the lowest collision frequency. Individual-policy PPO increases collisions by approximately 40\%, while No-Curriculum PPO significantly increases risk. Training directly in Stage~3 without curriculum results in substantially higher collision rates, demonstrating reduced stability under complex traffic conditions.

Min. Passing Distance (m): The proposed method maintains the largest safety margins, approximately 25\% higher than reward-ablation variants. No-Curriculum and Stage-isolated models tend to generate more aggressive trajectories with reduced safety buffers.

Episode Duration: The proposed policy completes navigation tasks efficiently while maintaining safety. Individual-policy PPO and Distance-only variants require up to 15\% longer durations, whereas Stage~3-only training shows the slowest convergence (exceeding 120 steps), indicating unstable decision-making under complex conditions.

Regulation Compliance: 
\begin{itemize}
  \item Head-On Right Turn Rate (Rule 14): Proposed method achieves substantially higher compliance compared to Stage~3-only training.
  \item Overtake Clearance Rate (Rule 13): Composite reward and proposed method maintain stable clearance margins, while Distance-only shows noticeable degradation.
  \item Crossing Right Yield Rate (Rule 15): Proposed approach maintains consistently higher compliance than non-composite reward variants and Stage~3-only training.
  \item Parallel Safe Distance Rate: Highest safety margins are maintained by proposed method and composite reward models, while other variants show reduced separation distances.
\end{itemize}

Taken together, the combination of shared recurrent policy learning, staged curriculum training, and composite reward design improves the robustness, safety, and reliability of onboard navigation under partial observability. These results further support the effectiveness of shared onboard policy learning for scalable and dependable autonomous navigation in IoT-enabled maritime environments.

\subsection{Behavioral Analysis and Generalization Insights}

\begin{figure}[t]
    \centering
    \includegraphics[width=0.8\linewidth]{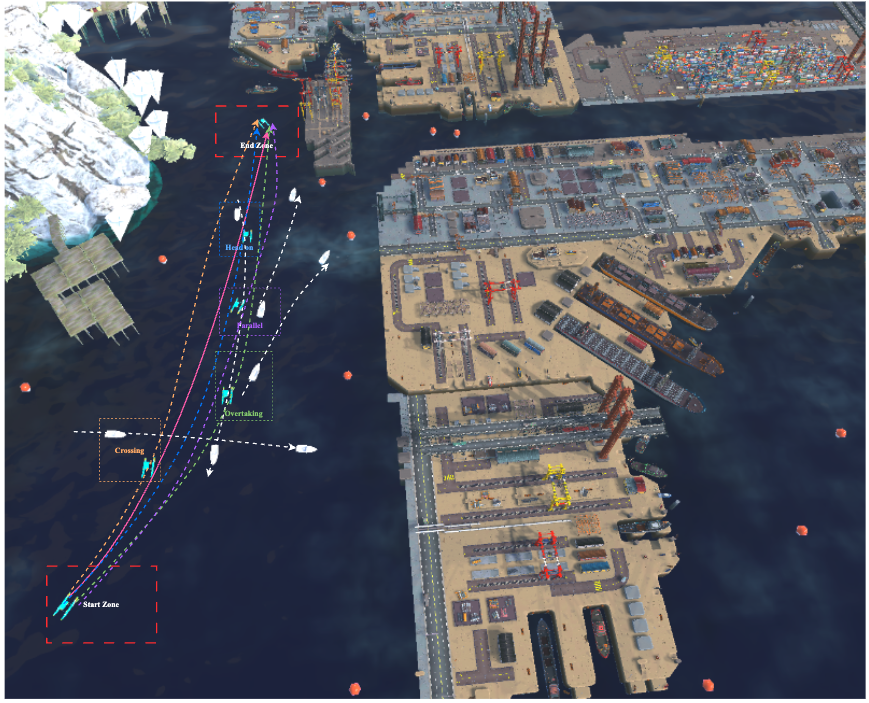}
    \caption{Representative onboard navigation trajectories under four COLREGs encounter conditions in the Stage~1 LA port scenario. Pink: baseline without interaction; light blue: ego USV; white: target vessels (ASVs); dashed arrows: ASV planned paths; solid lines: USV trajectories.}
    \label{fig:la-trajectories}
\end{figure}

To evaluate the effect of CL and the impact of the LSTM-based shared recurrent PPO architecture on agent behavior, we visualize representative trajectories from the Stage~1 (1vs4) LA port scenario (Fig.~\ref{fig:la-trajectories}). The USV navigates from the Start Zone (bottom-left, red dashed box) to the End Zone (top-center, red dashed box) under four COLREGs-defined encounters. Trajectory colors indicate interaction types, with the light blue vessel as the ego USV (Own Ship), white vessels as ASVs (Target Ships), white dashed arrows as ASV planned paths, and solid colored lines as USV motions. Only one start–end configuration is shown for clarity; actual experiments randomize both.

The pink line shows the baseline without interference. The orange trajectory (crossing, Rule~15) depicts the OS yielding to starboard when the TS is detected on its right front quarter, adjusting speed/heading to pass astern. The green trajectory (overtaking, Rule~13) shows the OS approaching from behind ($<22.5^\circ$) and maneuvering around the TS with a safe passing side. The blue trajectory (head-on, Rule~14) involves both vessels altering course to starboard at a $5$–$10^\circ$ approach. The purple trajectory (parallel) maintains lateral distance with minimal deviation. Stage~1’s low-traffic setting allows gradual acquisition of fundamental collision-avoidance behaviors, where parallel interactions cause minimal deviation, head-on and crossing require larger adjustments, and overtaking produces the most substantial detours.

\begin{figure}[t]
    \centering
    \begin{subfigure}[b]{0.8\linewidth}
        \centering
        \includegraphics[width=0.8\linewidth]{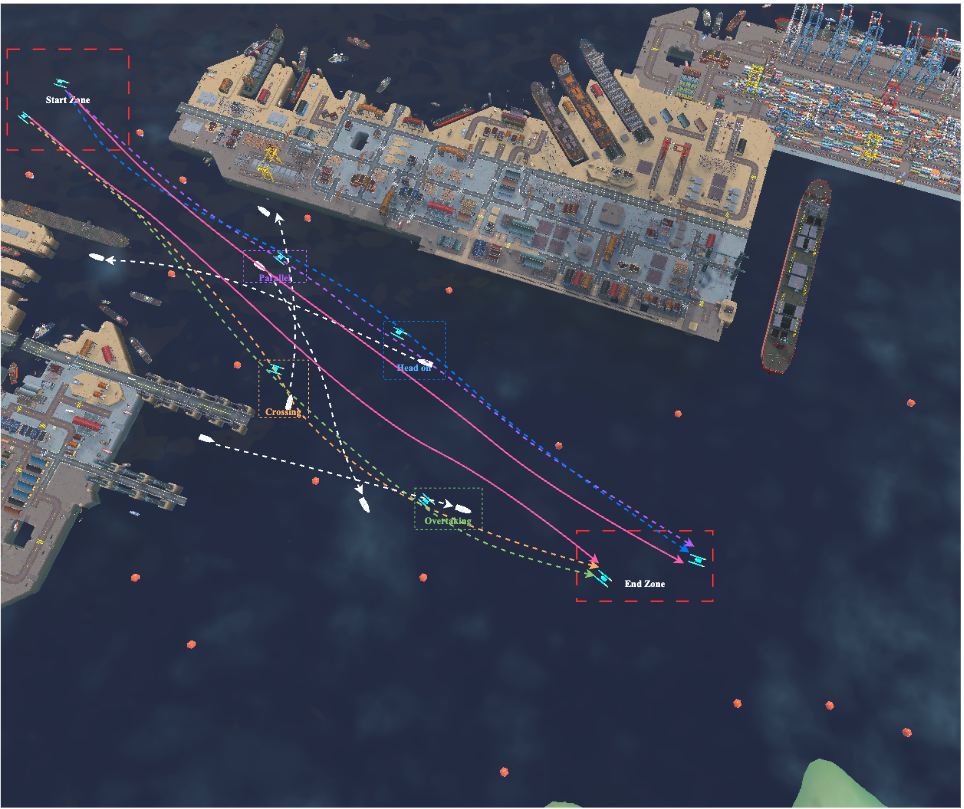}
        \caption{2vs2 scenario with moderate traffic, illustrating stable onboard navigation and COLREGs-compliant maneuvers.}
        \label{fig:sg2vs2}
    \end{subfigure}
    \hfill
    \begin{subfigure}[b]{0.8\linewidth}
        \centering
        \includegraphics[width=0.8\linewidth]{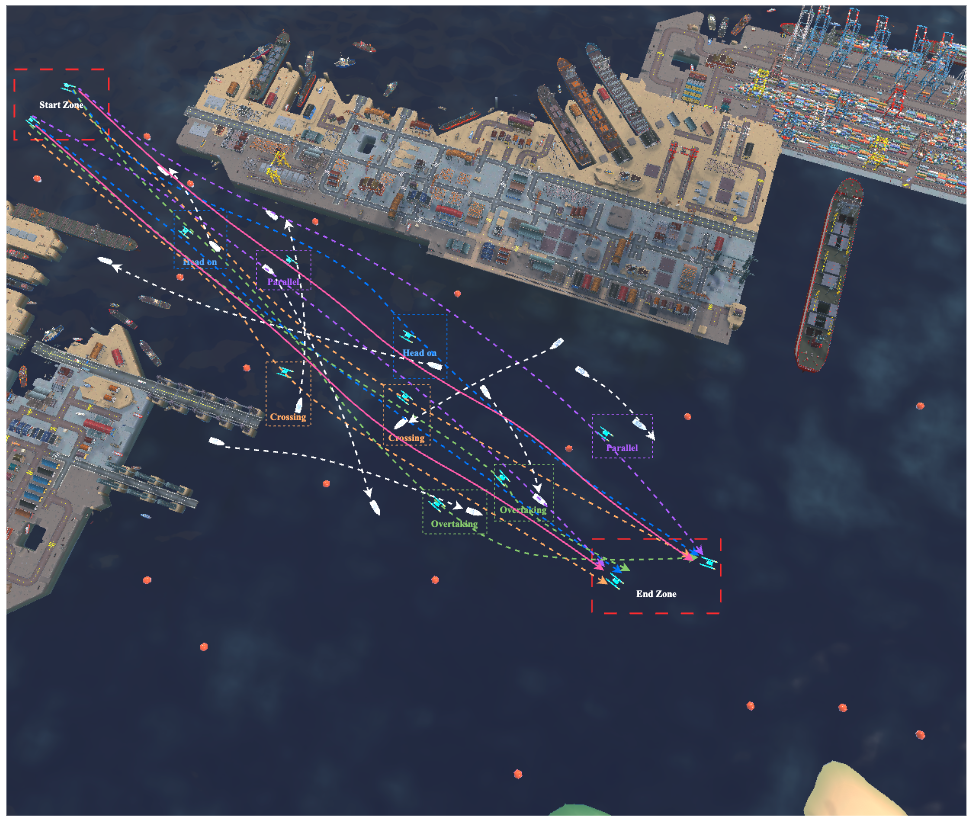}
        \caption{2vs4 dense-traffic scenario illustrating robust onboard navigation under dynamic multi-vessel interactions.}
        \label{fig:sg2vs4}
    \end{subfigure}
    \caption{Stage~2 onboard navigation trajectories under varying traffic densities, demonstrating stable COLREGs-compliant behavior and adaptive path adjustment.}
    \label{fig:sg-trajectories}
\end{figure}

Stage~2 increases complexity with two busy terminals, multiple buoys, stationary workboats, and dynamic ASVs selecting two distinct COLREGs encounters per episode. In 2vs2 (Fig.~\ref{fig:sg2vs2}), USVs generally maintain smooth paths, with overtaking resolved by starboard course changes and parallel encounters showing minimal deviation. Head-on maneuvers involve early right turns, and crossing cases prompt gradual mid-path adjustments. In 2vs4 (Fig.~\ref{fig:sg2vs4}), higher density leads to greater baseline deviations. Overtaking requires larger lateral shifts, crossing triggers earlier turns, head-on cases still resolve via rightward avoidance but with larger detours, and parallel encounters sometimes drift due to cumulative avoidance.

Comparisons show that traffic density consistently increases path deviation and prompts earlier avoidance in 2vs4. Overtaking produces the largest offsets, crossing is resolved earlier in denser traffic, and parallel/head-on remain structured but less precise.

\begin{figure}[t]
    \centering
    \begin{subfigure}[b]{0.8\linewidth}
        \centering
        \includegraphics[width=0.8\linewidth]{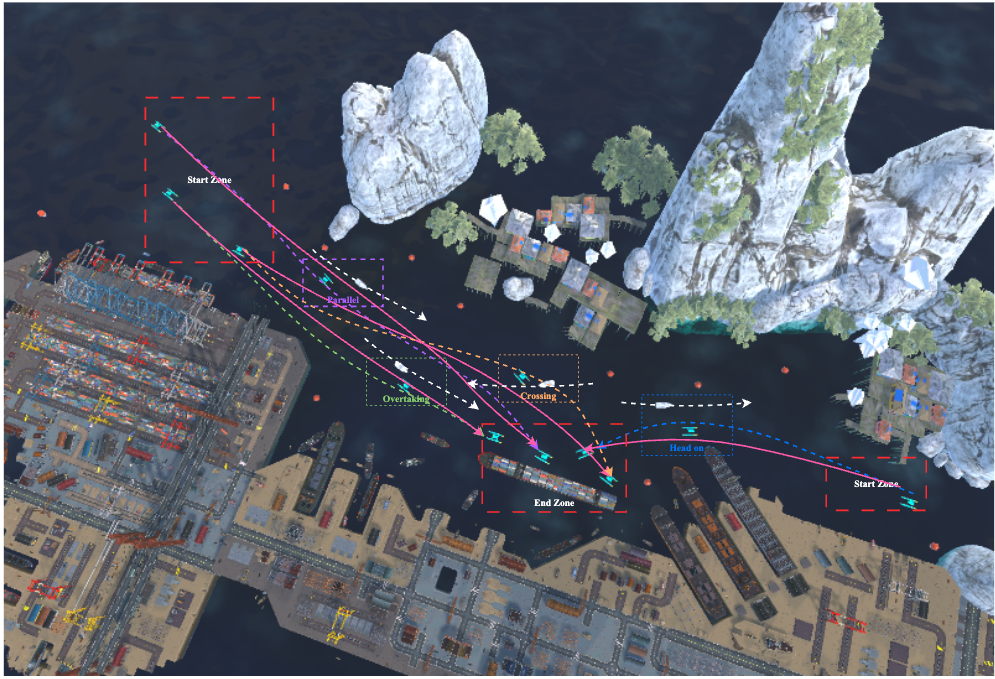}
        \caption{4vs4 scenario in the Rotterdam port, showing stable onboard navigation in narrow-channel conditions.}
        \label{fig:r-trajectories-rotterdam-4vs4}
    \end{subfigure}
    \hfill
    \begin{subfigure}[b]{0.8\linewidth}
        \centering
        \includegraphics[width=0.8\linewidth]{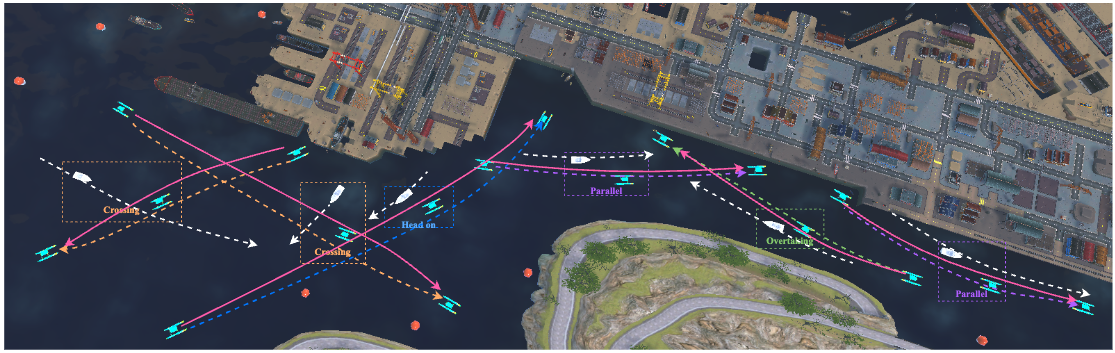}
        \caption{6vs6 scenario with increased multi-vessel interactions and environmental disturbances, illustrating adaptive onboard navigation behavior.}
        \label{fig:r-trajectories-rotterdam-6vs6}
    \end{subfigure}
    \begin{subfigure}[b]{0.8\linewidth}
        \centering
        \includegraphics[width=0.8\linewidth]{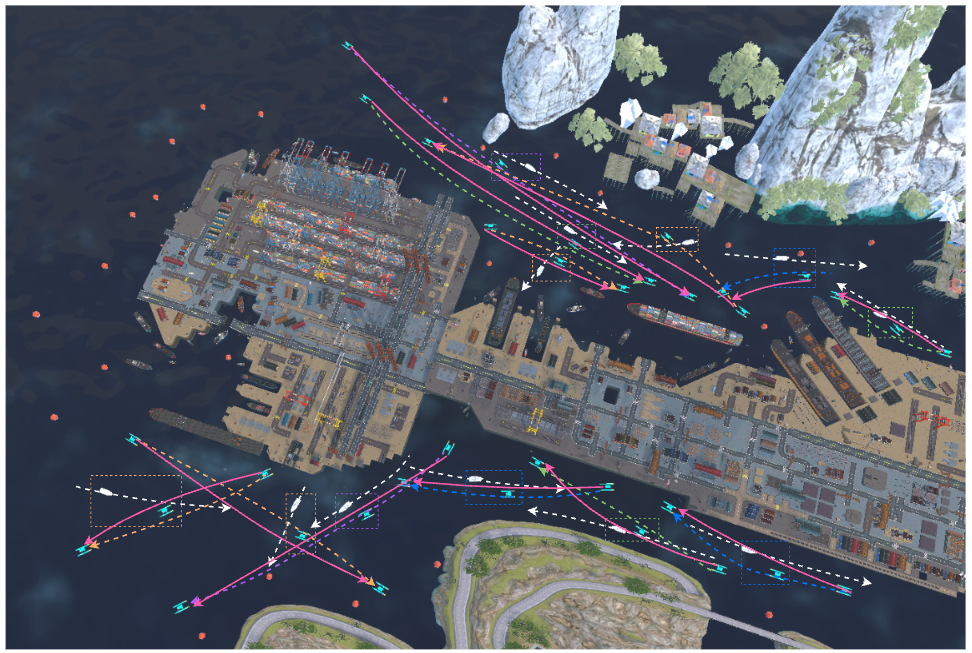}
        \caption{12vs12 high-density scenario demonstrating robust onboard navigation under complex and dynamic traffic conditions.}
        \label{fig:r-trajectories-rotterdam-12vs12}
    \end{subfigure}
    \caption{Stage~3 onboard navigation trajectories in the Rotterdam port. Narrow channels, dense interactions, and environmental disturbances require early avoidance and conservative navigation behavior.}
    \label{fig:r-trajectories-rotterdam}
\end{figure}

Stage~3 (4vs4, 6vs6, 12vs12) in Rotterdam adds narrow waterways, dense multi-vessel interactions, and wave/wind disturbances. In 4vs4 (Fig.~\ref{fig:r-trajectories-rotterdam-4vs4}), all four encounter types show early avoidance, with deviations shaped by narrow-channel constraints. In 6vs6 (Fig.~\ref{fig:r-trajectories-rotterdam-6vs6}), pre-emptive avoidance persists but trajectories align closer to baseline, indicating adaptability. Parallel nearly matches the baseline, and head-on/overtaking are smoother with maintained clearance. The 12vs12 case (Fig.~\ref{fig:r-trajectories-rotterdam-12vs12}) tests generalization in highly congested, disturbed waters. Agents anticipate conflicts early, maintain separation bands in multi-crossings, and preserve regulation compliance with reduced oscillations. Shared recurrent PPO with CL enables coordinated, temporally aware navigation robust to density and disturbances.

\begin{figure}[t]
    \centering
    \includegraphics[width=1.1\linewidth]{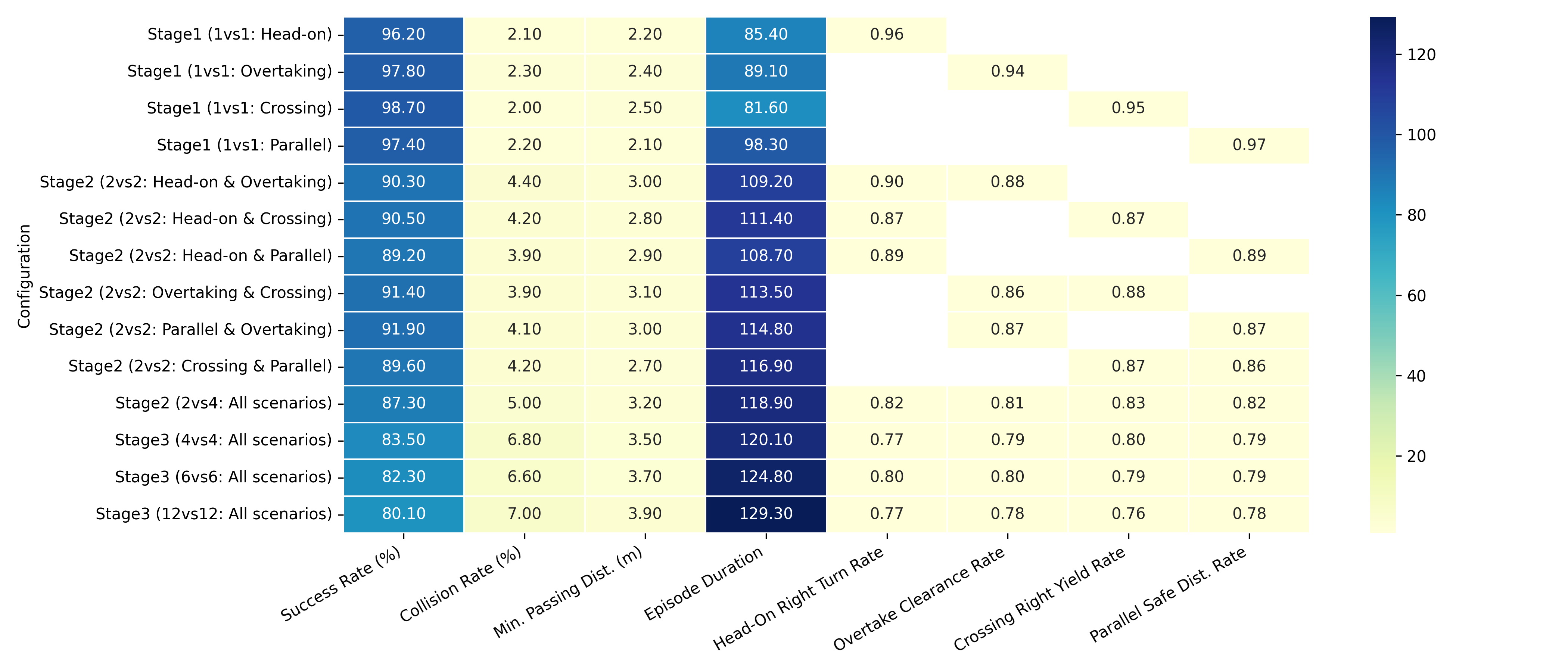}
    \caption{Heatmap of performance metrics across CL stages.}
    \label{fig:heatmap}
\end{figure}

Building on trajectory insights, aggregated metrics (Fig.~\ref{fig:heatmap}) assess robustness, safety, and rule compliance across stages. Stage~1 yields $>$96\% success with $<$2.5\% collisions and near-perfect regulation compliance. Stage~2 sustains $\sim$90\% success in 2vs2 but drops to 87.3\% in 2vs4, with increased collisions and longer episodes. Rule compliance also declines slightly under higher density. Stage~3 shows moderate success drop (83.5\% $\rightarrow$ 80.1\%) but improved minimum passing distances and consistent COLREGs adherence, reflecting stable, safe behavior under maximum complexity.

These observations provide several practical insights: (1) staged training in realistic port simulations supports progressive learning of safe navigation behaviors prior to deployment; (2) quantitative COLREGs-related metrics can serve as useful indicators for evaluating navigation reliability; (3) performance degradation under high-density traffic highlights the importance of adaptive safety margins and conservative maneuvering; and (4) curriculum-guided training combined with behavioral evaluation can improve the robustness and reliability of autonomous maritime navigation in complex operational environments.

\section{Conclusion}
\label{sec:6}

This paper presented a curriculum-guided DRL framework for reliable onboard navigation of IoT-enabled maritime autonomous surface vehicles in complex port environments. The proposed approach is built on a shared recurrent PPO architecture under a centralized training and decentralized execution paradigm. By incorporating temporal memory through LSTM and staged curriculum progression, the learned policy achieves stable, regulation-aware navigation under varying traffic densities and environmental disturbances.

Behavioral analysis shows that the learned policy develops consistent navigation strategies, including early conflict anticipation, smooth overtaking, structured head-on avoidance, and stable separation in multi-vessel interactions. These behaviors persist in progressively complex port scenarios and under environmental disturbances, indicating that the policy captures transferable navigation patterns rather than scenario-specific behaviors. Experimental results demonstrate stable performance across stages, maintaining high navigation success, low collision frequency, and consistent regulation-compliant behavior under increasing complexity.

The results highlight the effectiveness of shared recurrent policy learning combined with curriculum-guided training to improve the robustness, safety, and reliability of onboard navigation under partial observability. The use of quantitative regulation-related metrics provides interpretable indicators for evaluating navigation behavior, while staged simulation environments enable systematic assessment of policy robustness across varying operational conditions.

Future work will explore improving robustness through richer multi-modal sensing, including vision and radar fusion~\cite{kiran2021deep}, expanding scenario diversity to improve generalization across broader operational conditions~\cite{tobin2017domain,peng2018sim}, and investigating limited communication strategies for improved coordination in dense traffic~\cite{foerster2016learning, zhu2024survey}. Incorporating prior knowledge or imitation learning~\cite{hussein2017imitation, zhou2020smarts} can further improve training efficiency and safety in unfamiliar scenarios. In addition, hardware-in-the-loop and real-world validation will be investigated to assess real-time performance and deployment robustness, further bridging the gap between simulation and practical autonomous maritime systems.

\bibliographystyle{IEEEtran}
\bibliography{bibtex/bib/IEEEabrv,bibtex/bib/myrefs}

\end{document}